\documentclass[authoryear, preprint,12pt, 3p]{elsarticle}

\usepackage{amssymb}

\usepackage{xcolor}
\usepackage{caption}
\usepackage{subcaption}
\usepackage{amsmath,amssymb,amsfonts}
\usepackage{colortbl}
\usepackage[hyperfootnotes=true]{hyperref}
\usepackage{array}
\usepackage{float}
\usepackage{adjustbox}
\usepackage{multirow}
\usepackage[acronym]{glossaries}
\usepackage[english]{babel}
\usepackage{xcolor}

\biboptions{authoryear} 

\makeglossaries

\newacronym{COM}{COM}{Center-of-Mass}
\newacronym{HRI}{HRI}{Human-Robot Interaction}
\newacronym{MoCap}{MoCap}{Motion Capture}
\newacronym{IMU}{IMU}{Inertial Measurement Unit}
\newacronym{MARG}{MARG}{Magnetic, Angular Rate, and Gravity}
\newacronym{NWU}{NWU}{North-West-Up}
\newacronym{STS}{STS}{Sensor-to-Segment}
\newacronym{SDI}{SDI}{Strap-down Integration}
\newacronym{FPS}{FPS}{Frames per Second}
\newacronym{GT}{GT}{Ground-Truth}
\newacronym{FK}{FK}{Forward Kinematics}
\newacronym{EKF}{EKF}{Extended Kalman Filter}
\newacronym{KF}{KF}{Kalman Filter}
\newacronym{RMSE}{RMSE}{Root Mean Square Error}
\newacronym{DL}{DL}{Deep learning}
\newacronym{NN}{NN}{Neural Network}
\newacronym{RNN}{RNN}{Recurrent Neural Network}
\newacronym{biRNN}{biRNN}{bi-directional Recurrent Neural Network}
\newacronym{LSTM}{LSTM}{Long Short-Term Memory Network}
\newacronym{GRU}{GRU}{Gated Recurrent Unit}
\newacronym{TGCN}{TGCN}{Temporal Graph Convolutional Network}
\newacronym{CNN}{CNN}{Convolutional Neural Network}
\newacronym{GNN}{GNN}{Graph Neural Network}
\newacronym{NARX}{NARX}{Nonlinear Autoregressive Network with Exogenous Inputs}
\newacronym{GRNN}{GRNN}{General Regression Neural Network}
\newacronym{ReLU}{ReLU}{Rectified Linear Unit}
\newacronym{BN}{BN}{Batch Normalization}
\newacronym{MAE}{MAE}{Mean Absolute Error}
\newacronym{MSE}{MSE}{Mean Squared Error}
\newacronym{QAD}{QAD}{Quaternion Angle Distance}
\newacronym{PCO}{PCO}{Percentage of Correct Orientations}
\newacronym{PCK}{PCK}{Percentage of Correct Keypoints}
\newacronym{HPE}{HPE}{Human Pose Estimation}
\newacronym{ROS}{ROS}{Robot Operating System}
\newacronym{ONNX}{ONNX}{Open Neural Network Exchange}
\newacronym{API}{API}{Application Programming Interface}
\newacronym{SDK}{SDK}{Software development kit}
\newacronym{SotA}{SotA}{State-of-the-Art}

\begin{document}
\begin{frontmatter}

\title{Deep Inertial Pose: A deep learning approach for human pose estimation}

\author[inst1]{Sara M. Cerqueira \corref{cor1}}

\affiliation[inst1]{organization={Center for Microelectromechanical Systems},
            addressline={ University of Minho}, 
            city={Guimarães},
            postcode={4800 - 058},
            country={Portugal}}
\ead{id9484@uminho.pt, saracerqueira1996@gmail.com}

\author[inst1]{Manuel Palermo}
\ead{b12450@cmems.uminho.pt}
\author[inst1,inst2]{Cristina P. Santos}
\affiliation[inst2]{organization={LABBELS - Associate Laboratory},
            addressline={ University of Minho}, 
            city={Guimarães},
            postcode={4800 - 058},
            country={Portugal}}
\ead{cristina@dei.uminho.pt}

\cortext[cor1]{Corresponding author}

\begin{abstract}
Inertial-based Motion capture system has been attracting growing attention due to its wearability and unsconstrained use. However, accurate human joint estimation demands several complex and expertise demanding steps, which leads to expensive software such as the state-of-the-art MVN Awinda from Movella Technologies. This work aims to study the use of Neural Networks to abstract the complex biomechanical models and analytical
mathematics required for pose estimation. Thus, it presents a comparison of different Neural Network architectures and methodologies to understand how accurately these methods can estimate human pose, using both low cost(MPU9250) and high end (Mtw Awinda) Magnetic, Angular Rate, and Gravity (MARG) sensors. The most efficient method was the Hybrid LSTM-Madgwick detached, which achieved an Quaternion Angle distance error of 7.96º, using Mtw Awinda data. Also, an ablation study was conducted to study the impact of data augmentation, output representation, window size, loss function and magnetometer data on the pose estimation error. This work indicates that Neural Networks can be trained to estimate human pose, with results comparable to the state of the art fusion filters.
\end{abstract}





\end{frontmatter}


\section{Introduction}
\label{sec:intro}
In recent years, inertial-based \gls{MoCap} systems have been increasingly used for human motion tracking and analysis, in a wide range of applications. For example, in rehabilitation, this technology has been used in gait monitoring \cite{Sprager2015} and to support impairment diagnosis \cite{Luo2016}. In industry, they have been explored for ergonomic risk assessment \cite{MERINO201980}.  The main reason is the fact that they can be worn by the subject, in a natural and unconstrained manner, since they do not imply mobility restrictions. Also, they  can be used in both indoors and outdoors, without requiring a controlled environment on a restricted space, as camera-based solutions do \cite{Choe2019s2s}. 

This versatility potentiated the development of a considerable number of commercial inertial \gls{MoCap} systems, such as Rokoko’s Smartsuit Pro (Rokoko Electronics ApS, Denmark), Perception Neuron Pro (Noitom Ltd, Miami), Nansense BioMed (Nansens Biomed Inc., Los Angeles), Movella MVN Awinda (Movella, Netherlands), and Movella MVN Link (Movella, Netherlands). Movella Technologies is the leading company in 3D motion tracking. Movella MVN Awinda and  Link are considered a state-of-art system, being featured in several papers for different applications (e.g. research in HRC, ergonomics, sports) \cite{Gonzalez2021}. Also, they are one of the few inertial \gls{MoCap} systems validated for biomechanical analysis \cite{AlAmri2018xsenserror}. However, the high cost of these systems, derived not only from the expensive hardware, but also for the even more expensive licenses, required for the use of the property software for biomechanical analysis (MVN Analyze), pose barriers for their wide use on real and daily life applications. Although Movella’ systems can be used without its proprietary MVN Analyze software, this only provides raw data, without interpretation and whose processing requires strong technical and mathematical skills. 
Academia also followed this trend to develop \gls{IMU} based wearable technology. These wearables, however, tend to be custom-made and specific for an application. For example, single arm systems \cite{Ligorio2020, skulj2021} or specific for ergonomic assessment \cite{VIGNAIS2013}, not allowing general usage.

Human Pose Estimation (HPE), a specific application of human motion tracking, consists of estimating the position and orientation of human body keypoints, such as the joints, in a 2D or 3D space.
The estimation of Human joint orientation, from inertial data, i.e. 3D acceleration, 3D angular velocity, and 3D magnetic field measurements, is typically conducted with sensor fusion filters. This sensor fusion is performed to mitigate the drawbacks of each individual sensor. For example, the integration of the angular velocity of the gyroscope causes a significant orientation drift, due to the inherent sensor's measurement drift, even if small. The accelerometer, in turn, has a stable reference, provided by the gravity acceleration. However, it is typically a noisier sensor. At last, the magnetometer provides the plane heading by measuring the  earth's magnetic field direction. Withal, it is considerably affected by magnetic disturbances \cite{baldi2019}. Typical sensor fusion algorithms applied for angle estimation are the Kalman Filter \cite{Figueiredo2020} and its variants (e.g. Extended \gls{KF} \cite{Fan2018}, Unscented \gls{KF} \cite{Atrsaei2018}, Dual Linear \gls{KF} \cite{Ligorio2020}) and the complementary filters ( e.g. Madgwick's Gradient Descent filter\cite{Madgwick2011}, Mahony filter \cite{Mahony2008}, complementary filter \cite{Figueiredo2020}). The fusion of this inertial data results in the orientation of the sensor in a global coordinate system. Since the \gls{IMU} sensor is attached to a body segment, the segment orientation can be inferred from the sensor's orientation. However, this step requires a calibration step, also designated \gls{STS} alignment, where the relation between the relative orientation of the body segment coordinate system and the sensor coordinate system is defined. Then the relative orientation between segments, i.e, the joint angles, can be computed, towards pose estimation.  \gls{STS} calibration and sensor fusion are the most critical steps to achieve a high-accuracy system. In terms of \gls{STS}, the main source of error is the fact that it assumes a predefined static pose (or sequence of static poses) or a  dynamic movement sequence, easily leading to offsets due to mismatches between the calibration performed by the user and the reference, especially when fast/simple calibration procedures are necessary for real-world usage. As for the sensor fusion, it is mostly affected by the gyroscope integration, especially due to its accumulating bias drift, and the accelerometer/magnetometer reference vectors, due to biases and measurement noise.

Additionally, these steps are highly affected by wearability concerns, since sensor sliding can result in sensor-to-segment offsets. Therefore, they are still open research fields that have been receiving attention from investigators \cite{Ligorio2020,Pacher2020_ReviewSTS}.

The success and low error of Movella' MVN Awinda in human motion tracking and pose estimation is mainly due to its software, which comprises several differentiating steps that aim to mitigate state-of-art \glspl{IMU}-related sources of error. First, regarding gyroscope cumulative drift, besides their higher-quality hardware with a lower gyroscope bias characteristics, they also have a patented \gls{SDI} algorithm  that significantly reduces the gyroscope drift, which typically affects the estimation accuracy in the long term \cite{xsens_sdi}. Second, regarding sensor fusion, they developed a new \gls{KF}, entitled XKF3hm, that increases the orientation accuracy by minimizing the Earth magnetic field distortions and attaining a drift-free, absolute orientation \cite{Paulich2018mtw}. Third, regarding \gls{STS} calibration, they perform a dedicated and dynamic procedure, that does not rely in magnetometer data. This is an important step that ensures the alignment between the sensors and human body segments to allow an accurate estimation of the segment kinematics. Fourth, and the core element, they developed a biomechanical model of the human body \cite{RoetenbergPhD} that allows the combination of all motion trackers information, reducing the effects of magnetic distortions. \cite{XsensMVN}. All this advanced and experience demanding pipelines for signal processing explain the high cost of these licenses. Besides all that, Movella textiles/straps were improved towards minimal sensor slide, which reduces sensor-to segment offsets and, if placed correctly, soft-body artifacts resulting from muscle movement. However, this also results in a complex, time-demanding setup, that many times reduces exploitation by industry and/or rehabilitation settings.

The aim of this paper is to investigate a different approach for sensor fusion for pose estimation that dismisses complex biomechanical models and advanced and theoretical mathematical models for motion tracking, using \gls{MARG} sensors, and if possible compensate sources of error resulting from poor sensor dressing setup. It explores \gls{DL} algorithms to abstract from this analytical methods by following a data-driven approach. Moreover, it benchmarks the algorithms performance using both low-cost, highly available \gls{MARG} sensors, specifically the MPU9250, and high-quality, expensive MTwAwinda sensors. With this approach we aim to determine if \gls{NN}-based approaches can learn how to surpass hardware and sensor fixation limitations, common on low-cost, widely available hardware, and compare the method's performance on a hardware that is near of ideal(MTwAwinda).

If successful, this methodology could provide a scalable and accessible solution for real-time pose estimation, eliminating the need for complex biomechanical models, allowing developers working on wearable prototypes to integrate these algorithms without requiring an exhaustive understanding of the underlying technical complexities. As a result, they can direct their resources toward other critical aspects, such as hardware design, textiles, aesthetics, and overall prototype validation. Also, by testing these models with both low-cost and high-quality inertial sensors, this study provides a broader overview of its performance.

Joint angle estimation is a fundamental component of many advanced motion analysis systems, making this approach applicable across various fields such as rehabilitation, ergonomics, and sports. In rehabilitation, it could support personalized assessments and enable effective tracking of recovery progress without reliance on expensive equipment \cite{Neurorehabilitation}. In ergonomics, it could facilitate real-time monitoring of body posture \cite{MARTINS2024}, helping to prevent repetitive strain injuries and enhance workplace safety. In sports, it could provide valuable insights into athletes' movements \cite{sports}, aiding in technique optimization, performance improvement, and injury prevention.

\subsection{Related Work}

Pose Estimation is a vast research field with a good amount of literature on using \gls{DL} algorithms to detect and predict 3D position \cite{MEMe2022,BENGAMRA2021}. However, it is typically performed on images and videos, whose challenges mostly rely on image quality/noise/artifacts, very different from the challenges of estimating pose from inertial data. 
\gls{DL} algorithms have already been explored on inertial data, however to reconstruct 3D human body pose from a small set of sensors. For example, Huang et al. \cite{Huang2018dipimu} addressed real time reconstruction of Human pose using only 6 high-cost MTwAwinda sensors \cite{Paulich2018mtw} with a \gls{biRNN}, achieving a error of 15.85$^{\circ}\pm$12.87$^{\circ}$ and 17.54$^{\circ}\pm$6.49$^{\circ}$, depending on the dataset used. Yi et al. \cite{Yi2021TransPose} improved the method (achieving 12.93º ± 6.15º and 8.85 ± 4.82, respectively, for the same datasets) with a multistage pose estimation and a fusion-based global translation estimation, using both a \gls{biRNN} and a \gls{RNN}. Working on the same topic and with the same data \cite{Liao2023}, in turn, proposed a temporal convolutional encoder and human kinematics regression decoder \gls{NN} that  attained an angle error of 13.37º ±  5.35º, and 13.76º ± 5.21º, respectively.

In gait, sensor reduction has also been investigated using \glspl{NN} by \cite{Alcaraz21} and \cite{Kumano_2023}. \cite{Alcaraz21}  employed \glspl{LSTM}, \glspl{GRNN} and \glspl{NARX} to estimate sagittal lower limbs' joint angles using only one \gls{IMU}, placed on the foot. The \gls{LSTM} outperformed the remaining, achieving errors between 1.91º and 2.57º, in comparison to the \gls{GT} data. However, the \gls{GT} of this study was the joint angles output of a \gls{KF} fed by data from 4 \glspl{IMU}, placed on the right side at the foot, lower leg, upper leg, and pelvis. \cite{Kumano_2023}, in turn, presented a \glspl{LSTM}-based \gls{NN} designated IMU2Pose, which estimated 17 joint angles from 3 \gls{IMU}s positioned at the both heels and left wrist, obtaining a \gls{RMSE} lower than 12.28º. However, this approach was trained using synthetic \gls{IMU} data, generated from the \cite{AIST_dataset} dataset, whose data was captured using Vicon optical motion capture system. 

Notwithstanding, these approaches are not directly comparable with our goal, since they do not explore the potential of \gls{NN}s to perform sensor fusion and joint angle estimation, required to estimate human body pose.   
Methodologies such as combining \gls{NN} methods with classic sensor fusion methods, such as \gls{KF}, have also been tested by Coskun et. al. \cite{Coskun2017} for pose estimation from images. The proposed method results outperformed the standalone \gls{KF} and standalone \gls{LSTM}, achieving state-of-art performance and reducing the joint error of the used dataset in 13.8\%. This puts hybrid approaches as interesting solutions for pose estimation.

\section{Methods}
\label{sec:methods}

This section describes the methodology behind the development of the framework for Inertial-based Human Pose Estimation. Therefore, it presents the equipments used in this work to build a custom dataset, along with the pre-process steps required to prepare the data. This is followed by an overview of the model framework and the tested \gls{NN} architecture. Ends with the description of the post-processing steps.

\subsection{Materials}
Two inertial-based systems were used in this work, as follows:
\subsubsection{Ergowear}
Upper-body kinematics were acquired using a smart garment prototype, designated Ergowear  \cite{resende2021ergowear}. The system (Figure \ref{fig:acquisition_setup_ergowear}) is composed by 9 MPU-9250 \glspl{IMU} (InvenSense, San Jose, CA, USA), a low cost \gls{IMU} that combines a tri-axial accelerometer (± 16 g) and a tri-axial gyroscope (± 2000 º/s) and a tri-axial magnetometer(±4900 µT). The \glspl{IMU} were placed on the nape, shoulders, trunk (T4 and S1 levels),and right and left upper arms, forearms, and hands.

\subsubsection{Movella MTw Awinda} 
\begin{figure}[t]
    \centering
    \begin{subfigure}[b]{0.45\textwidth}
        \centering
        \includegraphics[height=5cm]{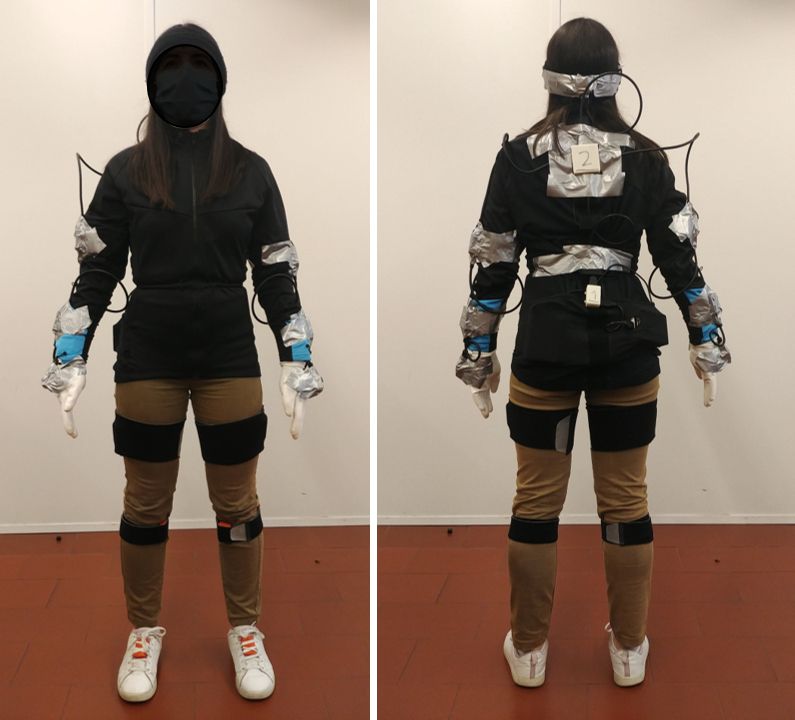}
        \caption{Ergowear}
        \label{fig:acquisition_setup_ergowear}
    \end{subfigure}
    \begin{subfigure}[b]{0.45\textwidth}
        \centering
        \includegraphics[height=5cm]{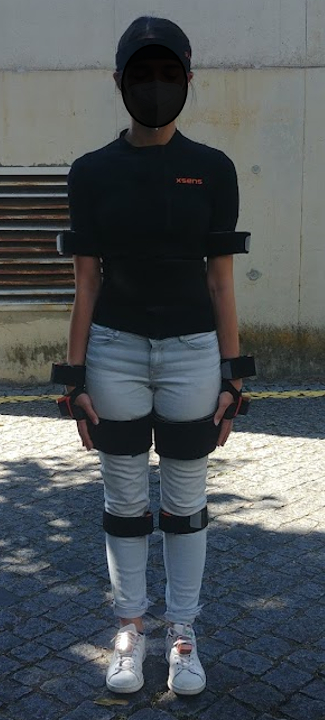}
        \caption{MTwAwinda}
        \label{fig:acquisition_setup_mtmanager}
    \end{subfigure}
    \caption{\textbf{(a)} Participant instrumented with Movella MTw Awinda and Ergowear. \textbf{(b)} Participant instrumented with two Movella MTw Awinda systems on top of each other.}
    \label{fig:acquisition_setup}
\end{figure}

Full body kinematics were acquired using the the Movella MTw Awinda inertial \gls{MoCap} system (Movella Technologies B.V., The Netherlands). It is composed of 17 wearable \gls{IMU} sensors, which communicate via wireless with a base station module, connected to a computer. This system was used in two manners: i) with the proprietary Movella MVN Analyze software, validated in \cite{AlAmri2018xsenserror}), to acquire \gls{GT} data. This software uses the Mtw trackers data to drive a biomechanical model of the subject that provides accurate positional and kinematic data, resulting in information on 23 human body segments. ii) with the free MTmanager software, to acquire high-quality raw sensor data. The system is depicted in figure \ref{fig:acquisition_setup_mtmanager}. 

\subsection{Complete Inertial Pose Dataset}

A custom dataset, which is already publicly available \cite{palermo2022cipdatabase}, was elaborated to develop and validate the inertial \gls{HPE} approaches. To fully explore the potential of these methods and evaluate its performance, two data acquisitions were performed, with very different characteristics; one using low cost hardware on a indoor environment full of magnetic interference, which might be found in rehabilitation, office or industrial settings; and the other using high-quality hardware (Mtw Awinda (Movella Technologies, B.V., The Netherlands), running MtManager software), on an outdoor environment, selected to decrease environment magnetic disturbances and, therefore, provide a best case scenario. Both data acquisitions followed a similar protocol in terms of movement variability.
Although both sensors (MPU9250 from Ergowear and the MTw motion tracker from  Mtw Awinda) have available firmware for sensor fusion that provide sensor's orientation, this output is not very stable for the low-cost sensors(MPU9250) and the programmer has no control over the implemented fusion filter. Therefore, it is not recommended in applications where high accuracy is required. As consequence only raw \gls{MARG} sensor data, i.e. 3D acceleration, 3D angular velocity, and 3D magnetic field measurements, were acquired. Both raw sensor data was related and syncronized with \gls{GT} data coming from the Movella MVN Analyse software system. A hardware trigger was used to start the acquisition on both systems.
These two data acquisitions were conducted aiming to verify if NN-based approaches can learn how to surpass hardware and sensor fixation limitations, common on low-cost, widely available hardware and benchmark against a hardware that is near of ideal. A more complete description of the datasets can be read on our complementary publication \cite{palermo2022complete, palermo2022cipdatabase} focusing on the data.
\subsection{Pipeline Development and Data Calibration}
\label{sec:dataset_preparation}

\begin{figure*}[t]
    \centering
    \includegraphics[width=0.99\textwidth]{ 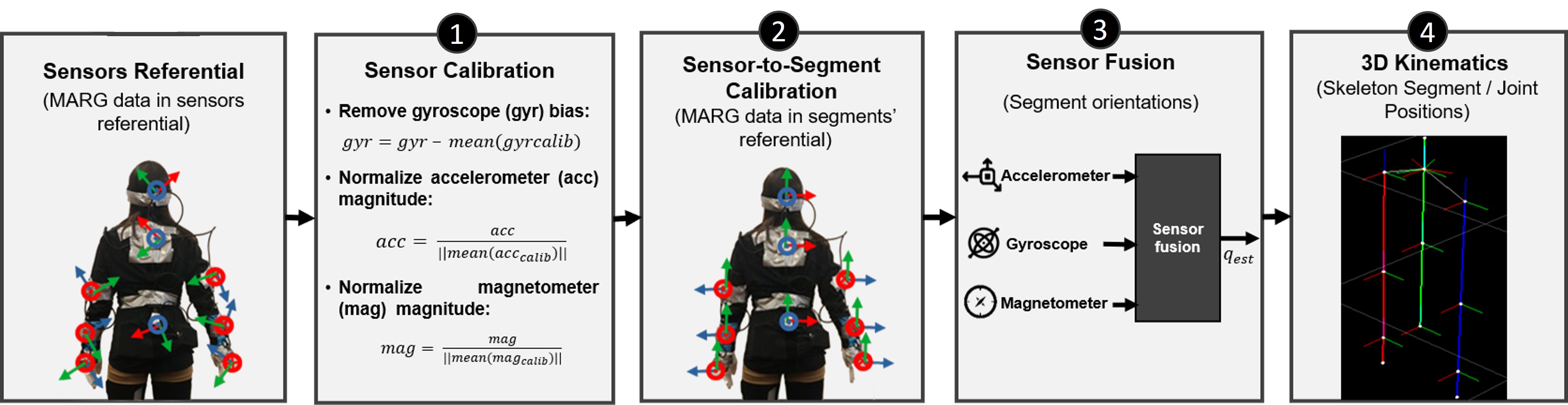}
    \caption{ Complete pipeline from raw sensor data to full body pose. }.
    \label{fig:methods_dataset_processing_steps}
\end{figure*}

A multi-step pipeline was implemented, compliant with inertial \gls{HPE} literature \cite{lopez2016imusreview, picerno2017review, Pacher2020_ReviewSTS} guidelines, to obtain the skeleton pose and compare results with the commercial \gls{MoCap} systems. Figure \ref{fig:methods_dataset_processing_steps} depicts this pipeline, whose steps are as follows:
\textbf{step 1)} sensors' scale/bias calibration 
\textbf{step 2)} sensor-to-segment calibration \cite{picerno2017review, Choe2019s2s}; 
\textbf{step 3)} whole body sensor-fusion;  Traditionally, this sensor fusion is performed for each sensor individually with classical filters (e.g. Kalman, Madgwick, among others) \cite{roetenberg2009xsens}. Innovation comes from the use of \gls{DL} methods to replace these traditional approaches for sensor fusion. 
\textbf{Step 4)} is related to the 3D kinematics tool developed to allow easy visualization of the estimated pose.

When using raw \gls{MARG} data from the higher-end hardware from Movella, however, the sensor bias  calibration step is not required, since the manufacture offers a sensor's self-calibration procedure.


\subsection{Sensor Fusion}
\label{sec:methods_sensor_fusion}

Calibrated \gls{MARG} sensor data, in the segments referential (after \gls{STS} calibration the sensor data is in the same referential as respective segments), was then fed to sensor-fusion methods (step 3 of figure \ref{fig:methods_dataset_processing_steps}). These calculate the orientation (represented as unit quaternions) of each segment. The orientation at the initial timestep is calculated using the TRIAD method \cite{markley2014ahrsfundamentals}, and is close to the unit quaternion since subjects will always start the trials in the same calibration pose, defined as reference.
 
\begin{figure}[ht]
    \centering
    \includegraphics[width=\textwidth]{ 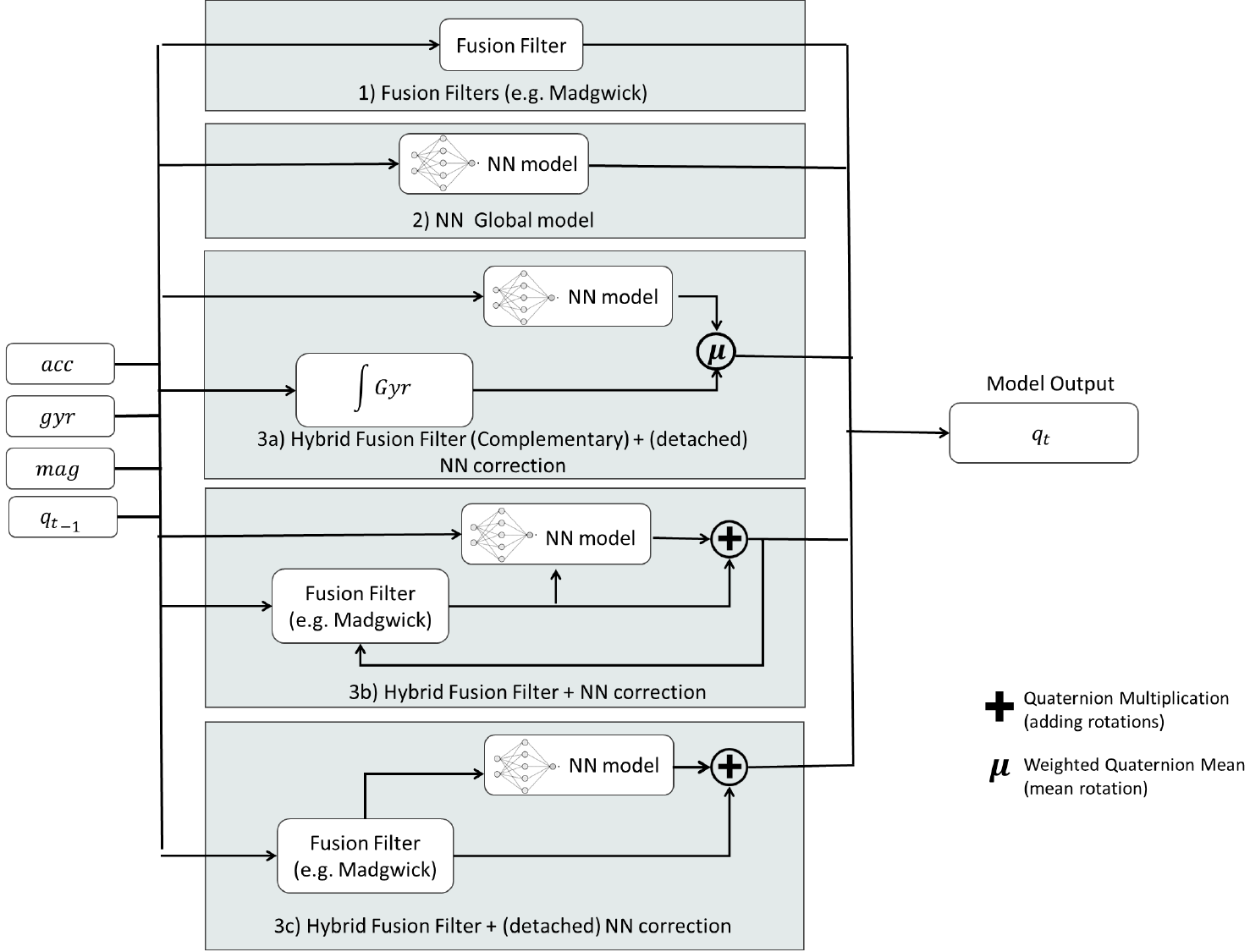}
    \caption{Multiple sensor-fusion architectures: 
    \textbf{1)} Classical Filter-fusion;
    \textbf{2)} Neural-Networks (global);
    \textbf{3a)} Neural-Networks (complementary approach, abbreviated as Hybrid-Complementary-NN);
    \textbf{3b and 3c)} Hybrid Filter-Neural-Networks, abbreviated as Hybrid-CFF-NN.
    }
    \label{fig:methods_inertial_pose_estimation_models}
\end{figure}

Three types of approaches with similar inputs and outputs were explored, as depicted in figure \ref{fig:methods_inertial_pose_estimation_models}:

\subsubsection{Classical Fusion-Filter}
\label{sec:methods_sensor_fusion_filters}

Follows the model-based classical literature \cite{markley2014ahrsfundamentals, Mahony2008, Madgwick2011, Sabatini2011EKForientation} and serves as a strong and relevant baseline on which most existing solutions are based \cite{roetenberg2009xsens, XsensMVN}. It relies on mathematical analysis of \gls{MARG} complementary sensor data (i.e. gyroscope to capture high frequency dynamics and accelerometer/magnetometer to provide global reference) \cite{markley2014ahrsfundamentals}. In simpler complementary methods \cite{Madgwick2011, Madgwick2020ECF, Mahony2008, markley2014ahrsfundamentals}, two different orientation solutions are averaged, by a gain factor: i) the gyroscope data integrated over time (dependent on previous step, making it recurrent), and ii) the accelerometer and magnetometer data using a TRIAD method (not dependent on previous steps). 

\subsubsection{Neural-Networks}
\label{sec:methods_sensor_fusion_nns}

Uses model-free \glspl{NN} to directly learn the mapping from \gls{MARG} sensor readings, to orientations from the training data (figure \ref{fig:methods_inertial_pose_estimation_models}-2).

This approach makes it straightforward to simultaneously combine data across all sensors, learning an internal latent biomechanical model representation, without explicit mathematical modeling. This combined with the use of information from multiple timesteps can reduce the effects of noise, magnetic disturbances and soft-body artifacts, which traditional systems have trouble dealing with, since these are hard to model explicitly \cite{Rong_2022}.

\subsubsection{Hybrid-Filter-Neural-Networks}
\label{sec:methods_sensor_fusion_filternns}

Combines model-based fusion-filters with model-free \glspl{NN}, with the objective of obtaining advantages from both approaches, namely: relying on model-based priors to improve generalization over the complete input space and facilitate training, while using \glspl{NN} to learn to address unmodeled system dynamics and perform whole body sensor fusion.

Two hybrid approaches for combining the \glspl{NN} with filters were explored:
\textbf{i)} the first (figure \ref{fig:methods_inertial_pose_estimation_models}-3a), abbreviated as Hybrid-Complementary-NN, is based on a complementary approach (inspired by the complementary filter). The gyroscope data is integrated across timesteps to produce the next quaternion, while a global quaternion is predicted by the \gls{NN}, based on the \gls{MARG} data. These are then added by a weighted importance factor.
\textbf{ii)} the second approach, abbreviated as Hybrid-CFF-NN uses a classical fusion-filter to output a quaternion prediction. Then, this data, along with the \gls{MARG} sensors readings, are fed into a \gls{NN}, to output a residual quaternion correction.
This approach can be divided in two more: 1) feedback-based approach (figure \ref{fig:methods_inertial_pose_estimation_models}-3b), where the quaternion from the \glspl{NN} is fed back into the filter; 2) a detached approach ( figure \ref{fig:methods_inertial_pose_estimation_models}-3c), where the filter's internal state is detached from the \gls{NN}. This means that the classical fusion-filter quaternion predictions are independent of the \gls{NN} predictions, contrary to the previous method.

\subsection{3D kinematics}
\label{sec:post_processing_temporal_filter}

The segment orientations can finally be combined with a kinematic model of the subject to obtain the 3D positions of the body segments. The kinematic model consisted of 9 segments for Ergowear (pelvis, T12, neck, right shoulder, left shoulder,right elbow, left elbow, right wrist and left wrist) and 17 segments for MTwAwinda (pelvis, stern, head, right shoulder, right upper arm, right forearm, right hand, left shoulder, left upper arm, left forearm, left hand,right upper leg, right lower leg, right foot,left upper leg, left lower leg, left foot). Default segments' lengths of the kinematic chain were defined based on \cite{Behrad2012} and the initial angles were set according to a N-pose. This way, the complete pose (orientation + position) is obtained, and can be used for easy visualization and evaluation (Figure \ref{fig:methods_dataset_processing_steps}, step 4). It should be noted that the kinematics are performed in global segment space instead of parent-child relative joints  as traditionally. This eases the process, since the filter fusion outputs are in the former representation.

\section{Experimental Protocol}
\label{sec:experimental}

This section details the experimental protocols, including details on the dataset creation, data pre/post-processing and networks training and evaluation. Variations of the network architectures are also considered at the end of the section.

\subsection{Dataset Details}
\label{sec:experimental_dataset}

21 (body mass = 66.0 ± 7.8 kg, body height = 171 ± 8.6 cm ) and 10 (body mass = 64.4 ± 8.5 kg, body height = 171 ±8.1 cm ) subjects,  participated in the Ergowear and MTwAwinda acquisitions, respectively. Multiple trials were collected with the participants performing 6 types of sequences (ranging from calibration, to daily-activities and random movements), each focusing on different movement dynamics and range of motions. Each sequence was repeated 3 (Ergowear) and 5 (MTwAwinda) times. This amounts to 2.5M and 1M samples of data sampled at 60Hz, synchronized with a \gls{GT} inertial \gls{MoCap} system. As aforementioned, more details on the data and movements performed can be found on our complementary publication \cite{palermo2022complete} and publicity available dataset \cite{palermo2022cipdatabase}.

Given the different number of sensors and, therefore, inputs / outputs contained in each dataset (9 vs 17), each one was treated separately to pre-process, train and evaluate solutions, following similar steps (Figure \ref{fig:methods_dataset_processing_steps}).

\subsubsection{Calibration}
\label{sec:experimental_dataset_calibration}

The low-cost Ergowear sensors were calibrated (step 1 of figure \ref{fig:methods_dataset_processing_steps}) to remove bias and scale factors over all axis, following standard procedures on \gls{MARG} sensor literature \cite{GONCALVES2021}.

Sensor-to-segment calibration (step 2 of figure \ref{fig:methods_dataset_processing_steps}) was then performed, transforming the inertial data from the sensors' referential to each of the corresponding segments' referential. The segments' referential aligns with the world referential in the \gls{NWU} frame of reference when the user holds the T-Pose. This is consistent with the frame of reference used by the \gls{GT}'s Movella Analyze software \cite{xsens_mvn_usermanual}, enabling direct comparison of the data.

For the MTwAwinda dataset, offline \gls{STS} calibration for each subject was performed using an optimization process \cite{nomura2020warmcmaes} to minimize the \gls{MSE} distance between sensor inertial data (accelerometer and gyroscope) and the respective \gls{GT} trajectories, yielding relatively low orientation offsets. This was performed using multiple long trials, presenting slow dynamics with high variability (mostly sequence and task trials), which resulted in lower error than the standard calibration trials. The transformation was assumed to be constant across trials for the same subject, since the sensors were firmly secured with straps with good grip.

Unfortunately, this approach did not work reliably for the Ergowear data, since there was some unpredictable sensor displacement across trials. This derived from the subject's movement and the fact that the smart-garment prototype was not capable to firmly secure the sensors to the subjects' segments. Instead, a simple static-calibration \cite{Choe2019s2s} method was used to obtain a reasonable transformation, using the static N-Pose data collected on the start of each trial (first 5 seconds). This method assumes a manual and fixed sensor placement, which is then corrected to the sensor's referential in the pitch/roll axis, using earth's gravity vector as reference.

\subsubsection{Pre-Processing}
\label{sec:experimental_dataset_preprocessing}

The datasets were then preprocessed\footnote{\url{https://github.com/ManuelPalermo/HumanInertialPose/tree/main/hipose/data/dataset_parsing}} to make the data more suitable, for training and evaluating the models. This included:
\textbf{i)} interpolation of missing or corrupt data samples;
\textbf{ii)} resampling of the Ergowear data to a constant 60Hz in order to match the \gls{GT} system (not required for the MTwAwinda since data was acquired at this frequency);
\textbf{iii)} filtering out all trials where issues were detected (\textit{e.g.} loose sensors, data desynchronization or corruption, etc...);
\textbf{iv)} removing the initial seconds ($\sim$5s) of each trial, which contained static N-Pose calibration data;
\textbf{v)} mapping all \gls{GT} segments' sensor data to the corresponding in each dataset (\textit{i.e} 9 upper-body segments for Ergowear and 17 full-body segments for the MTwAwinda, from the original 23 from the Movella\textsuperscript{TM} Analyze software);
%
%
\textbf{vi)} normalization of the data to more suitable ranges (accel in $m/s^2$ was converted to $g$, magnetometer was normalized by the average magnitude of the measured magnetic field, the gyroscope data in $rads/s$ already had reasonable values, as well as the unit quaternions used to represent the \gls{GT} orientations).

\begin{figure*}[t]
    \centering
    \includegraphics[width=0.99\textwidth]{ 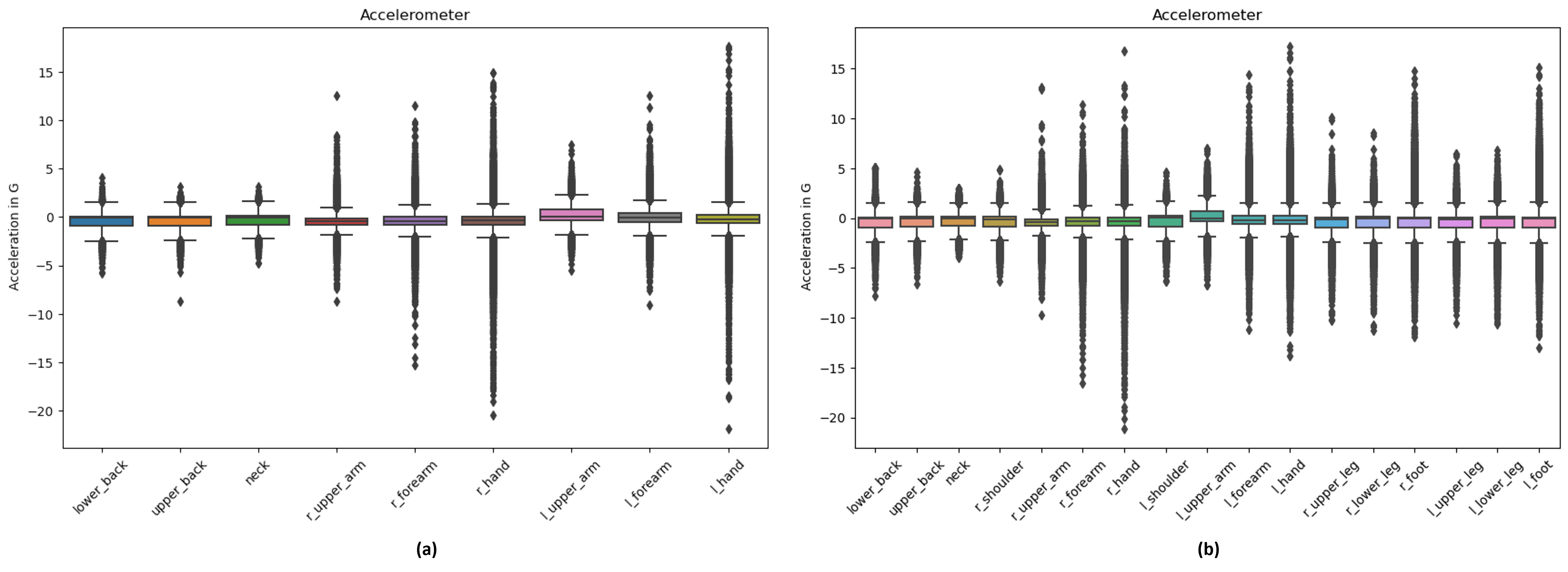}
    \caption{ Boxplots of the measured acceleration values per segment: (a) Ergoaware dataset, (b) MtwAwinda dataset. }.
    \label{fig:methods_acc_boxplot}
\end{figure*}

Table \ref{tab:descriptive_analysis} presents  the descriptive analysis of the two datasets, Ergoaware and MTwAwinda. As it can be observed, except for the magnetometer sensors of the MTwAwinda sensor, the wide range between the minimum and the maximum values, as well as the high and positive kurtosis, suggest that some values deviate substantially from the mean of the data. Also, one can observe that the statistics values are similar when comparing each sensor of each dataset, i.e. accelerometer with accelerometer, gyroscope with gyroscope. The main difference is verified between the magnetometers, which have similar values of mean and standard deviation. However, the maximum and minimum values are very different, possibly due to the high presence of magnetic perturbations in Ergoaware environment, as described in section \ref{sec:experimental_dataset}. Additionally, in terms of mean value, data is approximately centered around zero, and the standard deviation is around 0.6 G (6$m/s^2$) for the accelerometers, which is in agreement with the normal acceleration values for daily activities such as human walking \cite{acc_range} and 1.4 º/s for the gyroscopes. Figure \ref{fig:methods_acc_boxplot} presents the boxplot of all accelerometer sensors for Ergoaware and MTwAwinda datasets. The remaining sensors are not shown for space and readability reasons. Limbs extremities such as hands, foots, as well as the forearms, present a higher acceleration values range.

\begin{table*}[t]
    \centering
    \caption{Descriptive analysis of MTwAwinda and Ergoaware datasets}
    
    \maxsizebox{\textwidth}{!}{
    \begin{tabular}{l | c | c | c | c | c | c | c }
        Dataset                    & Sensor & Mean   & Standard Deviation  & Min     & Max     & Kurtosis  & Skewness\\
        \hline \hline
        \multirow{3}{*}{Ergowear}  & Acc    & -0.224  & 0.625   & -21.878 & 17.614  & 4.303 & 0.134\\
                                   & Gyr    & -0.009  & 1.181   & -37.392 & 34.139  & 16.09 & -0.087\\
                                   & Mag    & -0.093  & 0.611   & -37.651    & 50.417       &9.839       & 0.173 \\
        \hline
        \multirow{3}{*}{MTwAwinda} & Acc    & -0.261   & 0.63   & -21.145 & 17.211 & 5.798   & -0.014\\
                                   & Gyr    & -0.008   & 1.49   & -41.237 & 36.478 & 23.033  & -0.287\\
                                   & Mag    & -0.236   & 0.449  & -2.08   & 2.348  & -0.871  & 0.509\\
        \hline
    \end{tabular}%
    }
    \label{tab:descriptive_analysis}
\end{table*}

\subsubsection{Data Splits}
\label{sec:experimental_dataset_splits}

Each of the datasets was split by subjects into train (70\%), evaluation(20\%), and test(10\%). The split was kept constant to correctly perform benchmarking across multiple models and hyper-parameters. The training split only contained \textit{task}, \textit{circuit}, \textit{sequence} and \textit{random} trials, given that the \textit{calibration} and \textit{validation} trials contained little whole-body dynamic movements (these were still used in the validation and test splits). The resulting data is summarized on Table \ref{tab:dataset_splits}.

\begin{table}[ht]
    \centering
    \caption{Summary of the train, validation and test data splits for Ergowear and MTwAwinda datasets.}
    \maxsizebox{\textwidth}{!}{
    \small
    \begin{tabular}{l | c | c | c | c }
        Dataset                    & Splits & Subjects & Trials & Samples \\
        \hline \hline
        \multirow{3}{*}{Ergowear}  & train  & 15       & 184    & 773.6k  \\
                                   & val    & 4        & 24     & 101.9k  \\
                                   & test   & 2        & 51     & 236.4k  \\
        \hline
        \multirow{3}{*}{MTwAwinda} & train  & 7        & 117    & 480.1k  \\
                                   & val    & 1        & 20     & 73.2k   \\
                                   & test   & 2        & 41     & 188.5k  \\
        \hline
    \end{tabular}
    }
    \label{tab:dataset_splits}
\end{table}


\subsection{Sensor Fusion Models: Pipeline Details}
\label{sec:experimental_model}

All models were created and trained using the Pytorch \cite{Paszke2019pytorch} \gls{DL} library and with the help of Pytorch-Lightning \cite{William2019Lightning} to facilitate training and evaluation details, on a Python environment. 

The baseline filter-fusion models followed the same pipeline as the \gls{DL} models, but the training steps were skipped since these had no learnable parameters. Thus, only testing was performed by computing performance metrics on the same test set. All \gls{MARG} sensor data was used as input by default and all models produced orientations for all segments.

Full trajectories were used to validate and test the models. This is necessary, given the recurrent nature of the fusion-filters and the \gls{DL} models used (based on \glspl{RNN}), emulating usage in real applications. The full trajectories were also used for training the \gls{DL} models, performing backpropagation through time.

\subsubsection{NN models}
\label{sec:experimental_model_nns}
Following existing literature \cite{Huang2018dipimu, Yi2021TransPose, weber2020nnfusion}, recurrent based \glspl{NN} were used, with \gls{LSTM} being chosen as the default choice. These were preferred over window based approaches \cite{bai2018temporalcnns, vaswani2017attention} given the similar recurrent nature of classical filters \cite{markley2014ahrsfundamentals}.

\subsubsection{Hyperparameters}
\label{sec:experimental_model_hyperparams}
Reasonable hyper-parameters for training were found empirically and kept constant for all the combinations of models tried. The AdamW \cite{loshchilov2017adamW} optimizer was selected with an initial learning rate of 2e-3 which was decayed to 1e-5 over 25 epochs using a cosine-annealing schedule with warmup \cite{smith2019super} with a batch size of 64. Gradient clipping \cite{pascanu2013difficulty} with a range of [-0.2, 0.2] was applied during training for all models. At the end of the training, the best model was selected according to the validation loss. All layers, except for the outputs, were followed by \gls{BN} \cite{ioffe2015batchnorm} and a \gls{ReLU} non-linearity \cite{Glorot2011relu}.

\subsubsection{Data Augmentation and Regularization}
\label{sec:experimental_model_regularization}
To increase variability and, consequently, the model's capacity to generalize, random train-time data augmentation was applied to input and \gls{GT} data. This included: sensor gaussian or uniform noise; sensor dropout, i.e., random sensors' data is overwritten to zero; time-step dropout, i.e., time samples are removed; and small sensor data rotation offsets.
The strength of the augmentations was increased incrementally over each epoch. Additional train-time regularization was applied to most layers of the model in the form of dropout \cite{srivastava2014dropout}, with percentage of 20\% for all layers and weight decay parameter \cite{Krogh1992wdecay} with a value of 1e-5.

\subsubsection{Loss}
\label{sec:experimental_model_loss}
All \gls{DL} models were trained using the \gls{QAD} loss (Eq. \ref{eq:loss_qad}) \cite{huynh2009qad}, which directly minimizes the angle distance between two quaternions, while avoiding discontinuities associated with Euler representations and classical distance metrics (e.g. \gls{MSE}), which, when tried, led to high training instability.

\begin{equation}
    \label{eq:loss_qad}
    QAD = 2 \times \arccos(|q_{target} \cdot q_{pred}|), \quad QAD \in [0, \pi] 
\end{equation}

\subsection{Performance Metrics}
\label{sec:experimental_metrics}

All models (classical filters and \glspl{NN}) were evaluate based on the \gls{QAD} \cite{huynh2009qad} metric (Eq. \ref{eq:loss_qad}), converted to degrees for easier interpretation.

\textit{{\gls{QAD}}} - The average \gls{QAD} metric across time and across all sensors was used, in general, to compare results from different methods.

\textit{{QAD Boxplots}} - The \gls{QAD} distribution was plotted for each of the predicted segment orientations and visualized in boxplots, to better analyze the error for each individually.

\textit{{QAD Timeplots}} - The \gls{QAD} was plotted over time for each of the predicted segment orientations, giving a better notion of how the error evolves overtime (\textit{e.g.} drift) or in specific complex dynamics.

\textit{Inference Time} - Inference time was evaluated on an Intel\textsuperscript{TM} i9-10940X (3.30 GHz) CPU. This is crucial for assessing whether the application meets real-time requirements and refers to the time needed to load and process the inputs into the device and perform a forward pass for a single sample. The results were obtained by running inference on all samples in the test split.

\subsection{Model Variants}
\label{sec:model_variants}

Some alternative methods were tried, with regards to sensor fusion, some of which had been previously tried in the literature with varying degrees of success, these included:
\textbf{i)} the use of a magnetic rejection heuristic based on the measured magnetic field magnitude, similarly to Madgwick \textit{et. al.} \cite{Madgwick2020ECF};
\textbf{ii)} sensor fusion with or without the use of magnetometer data;
\textbf{iii)} different types of \glspl{RNN}, namely: vanila \gls{RNN}, \gls{GRU} \cite{cho2014gru}, \gls{TGCN} \cite{zhao2019tgcn} and \gls{LSTM} \cite{hochreiter1997lstm};
\textbf{iv)} different types of \gls{NN} output representations, namely: Euler, quaternion and 6D representation \cite{Zhou2019repr6d} (the representation is still converted to quaternion, as a post-processing step to compute the loss and metrics).

\textbf{v)} model training with temporal windows (1, 2, 5, and 50 steps) \cite{hochreiter1997lstm}, as opposed to the full trajectories. Although this approach is common in the literature, it was not performed by default as it lead to high error accumulation on the integration step of complementary methods.

\textbf{vi)} different loss functions: \gls{MSE}, \gls{MSE} but taking into account the shortest rotation arch \textit{qdist} and root-segment-relative \gls{QAD} (\textit{RelQAD}).

\section{Results}
\label{sec:results}
\subsection{Models' performance}

\subsubsection{Experiments with Neural Networks}

\begin{table*}[t]
    \centering
    \caption{Sensor Fusion average results, in QADº, and inference time (ms) on Ergowear and MTwAwinda data.}
    \maxsizebox{\textwidth}{!}{
    \begin{tabular}{l|l|c c|c c}
        Experiment                                 & Variant               & \multicolumn{2}{c|}{Ergowear}                  & \multicolumn{2}{c}{MTwAwinda}   \\ \cline{3-6}
                                                   &             & Ergowear QADº      & Inference time (ms)     & MTwAwinda QADº      & Inference time (ms)     \\ \hline \hline
        \multirow{3}{*}{Model-Free-NN}              & GRU              &  22.59  $\pm$ 15.73  & 0.30 $\pm$ 0.01    & 28.07  $\pm$ 23.98  & 0.33 $\pm$ 0.03    \\
                                                    & LSTM             &  21.57  $\pm$ 14.97  & 0.32 $\pm$ 0.01   & 27.97  $\pm$ 24.05  & 0.34 $\pm$ 0.02    \\
                                                    & TGCN             &  29.20  $\pm$ 21.43  & 0.50 $\pm$ 0.01   & 37.21  $\pm$ 34.13  & 0.60 $\pm$ 0.02    \\ \hline \hline
        \multirow{3}{*}{Hybrid-Complementary-NN}    & GRU              &  21.48  $\pm$ 17.87  & 0.43 $\pm$ 0.01   & 24.80  $\pm$ 23.15  & 0.47 $\pm$ 0.01    \\
                                                    & LSTM             &  21.58  $\pm$ 18.89  & 0.46 $\pm$ 0.01   & 24.30  $\pm$ 22.63  & 0.49 $\pm$ 0.02    \\
                                                    & TGCN             &  32.59  $\pm$ 23.60  & 0.63 $\pm$ 0.01   & 38.68  $\pm$ 36.13  & 0.74 $\pm$ 0.02    \\ \hline \hline
        \multirow{4}{*}{Hybrid-CFF-NN}              & LSTM-MG-detach   &  20.63  $\pm$ 16.45  & 1.64 $\pm$ 0.02   & 8.08   $\pm$ 6.67   & 1.63 $\pm$ 0.03    \\
                                                    & LSTM-MG-feedback &  diverged            & -                 & diverged            &  -  \\
                                                    & TGCN-MG-detach   &  21.89  $\pm$ 17.00  & 1.84 $\pm$ 0.02   & 7.96   $\pm$ 6.72   & 1.85 $\pm$ 0.01   \\
                                                    & TGCN-MG-feedback &  diverged            & -                 & diverged            &  - \\ \hline \hline
    \end{tabular}
    }
    \label{tab:table_NN_results}
\end{table*}

Table \ref{tab:table_NN_results} depicts the results from the 3 approaches, i.e., model-free \glspl{NN}, Hybrid-Complementary-\glspl{NN}, and Hybrid-Classic-Filter-\glspl{NN}, on both Ergowear and MTwAwinda data.

The best performance was achieved by the TGCN-MG-feedback, on the MTw Awinda dataset, attaining a \gls{QAD} error of 7.96º ± 6.72. However, when testing on the Ergowear dataset, the architecture with the best results was the LSTM-MG-detached, with a \gls{QAD} of 20.63º ± 16.45. Overall, the Hybrid-Classic-Filter-\glspl{NN}  outperformed the model-free-\glspl{NN} and Hybrid-Complementary-\glspl{NN}. Notwithstanding, when comparing between datasets, it is possible to observe that the results for Ergowear were more precise (meaning that result values are closer), ranging from 20.63º ± 16.45 to 32.59º ± 23.60 in contrast to 7.96º ± 6.72 to 38.68º ± 36.13 from MTwAwinda. In specific, the Model-free-\glspl{NN} performed better on the Ergowear than on MTwAwinda data. This is also supported by the standard deviation which is higher on the MTwAwinda data.  
Regarding inference time, as expected, it increases with the complexity of the model. Model-free \glspl{NN} had the lowest inference times (ranging from 0.3 to 0.6 ms), followed closely by the Hybrid-Complementary-\glspl{NN}. The Hybrid CFF-\glspl{NN}, due to the added complexity of combining a classic filter with \glspl{NN}, resulted in longer inference times (1.64 to 1.85 ms) .

\begin{figure*}[htbp]
    \centering
    \begin{subfigure}[b]{0.48\textwidth}
        \centering
        \includegraphics[width=\textwidth]{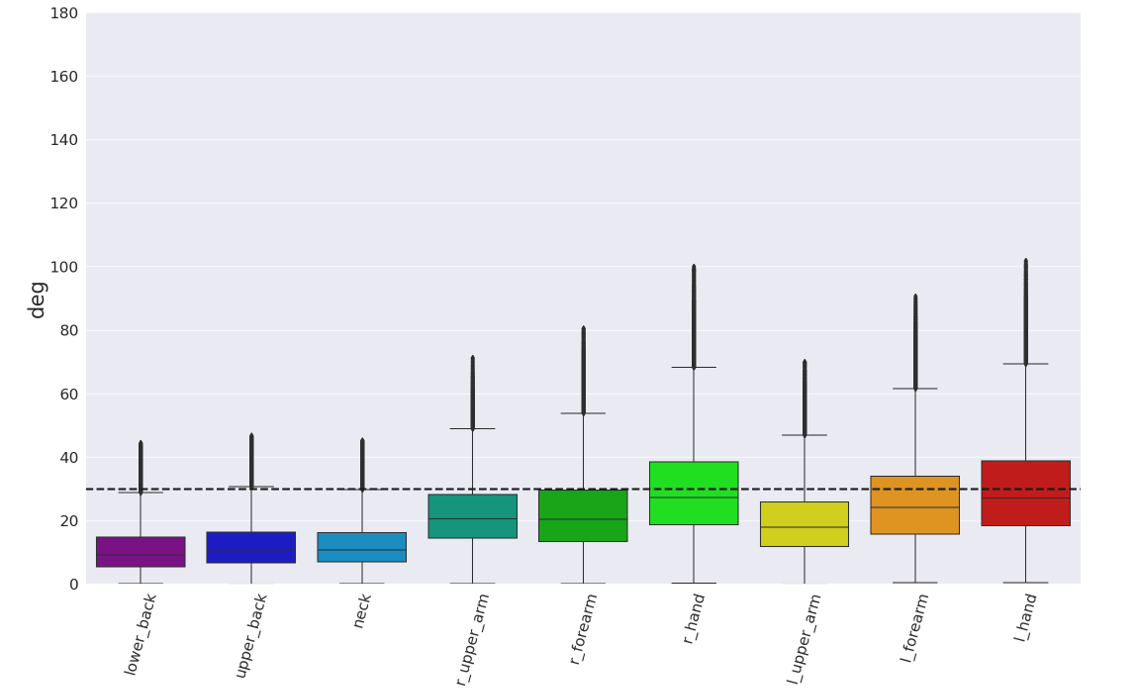}
        \caption{Ergowear's Model-free LSTM}
        \label{fig:results_ERGO_LSTM_GLOBAL}
    \end{subfigure}
    \hfill
    \begin{subfigure}[b]{0.48\textwidth}
        \centering
        \includegraphics[width=\textwidth]{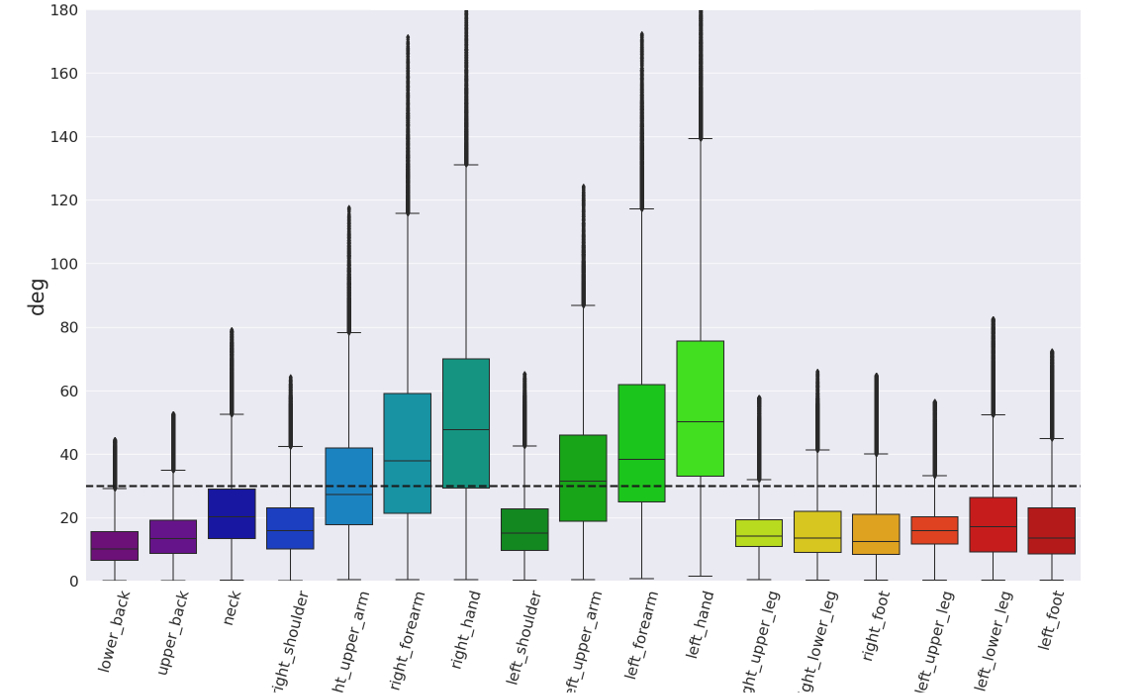}
        \caption{MTwAwinda's Model-free LSTM}
        \label{fig:results_MTGM_LSTM_GLOBAL}
    \end{subfigure}
    \vspace{0.5cm}
    \begin{subfigure}[b]{0.48\textwidth}
        \centering
        \includegraphics[width=\textwidth]{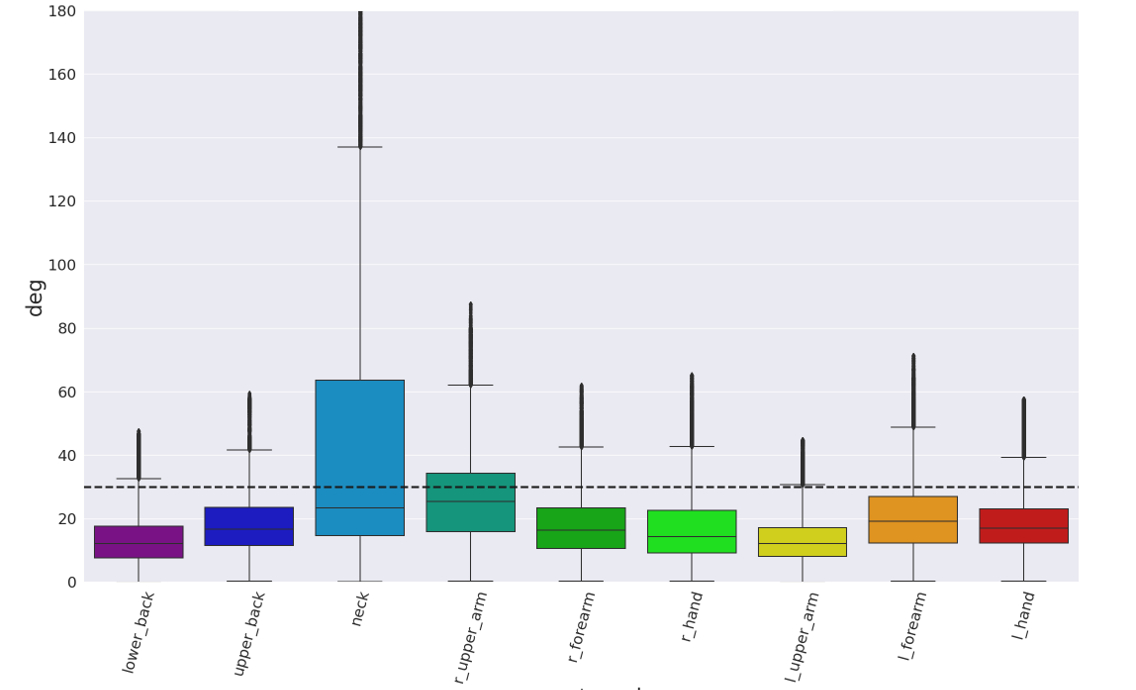}
        \caption{Ergowear's LSTM-MG-detach}
        \label{fig:results_ERGO_LSTM_MG_DETACH}
    \end{subfigure}
    \hfill
    \begin{subfigure}[b]{0.48\textwidth}
        \centering
        \includegraphics[width=\textwidth]{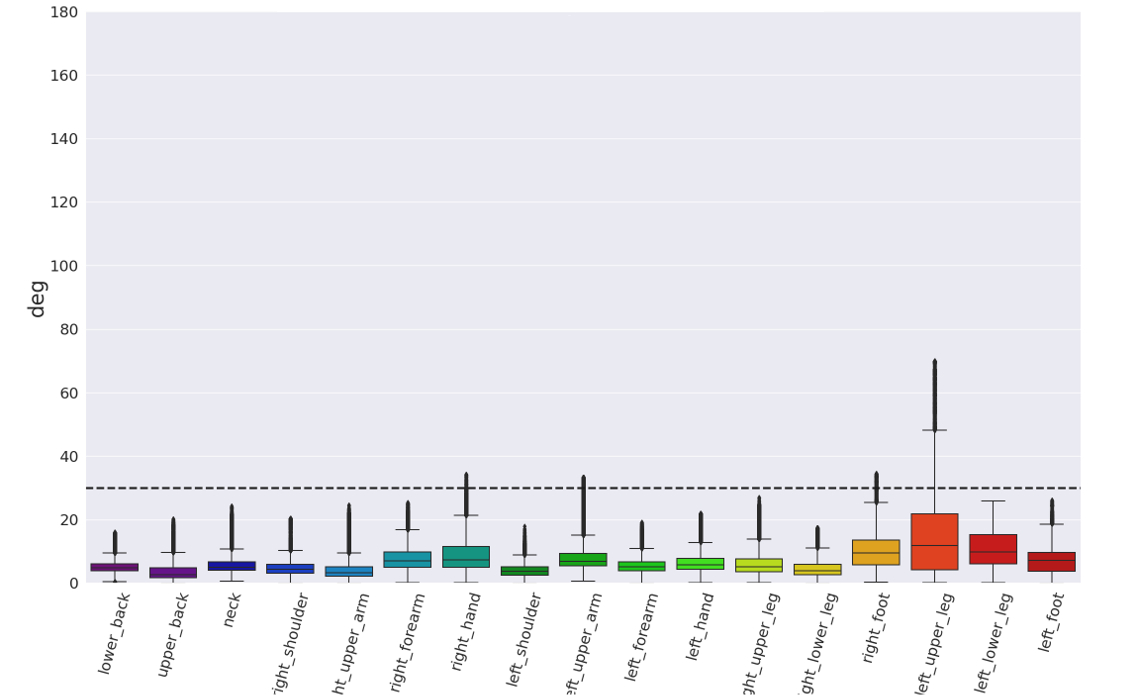}
        \caption{MTwAwinda's LSTM-MG-detach}
        \label{fig:results_MTGM_LSTM_MG_DETACH}
    \end{subfigure}
    \caption{Boxplots showing the error per segment: 
    (\textbf{a}) Model-free LSTM on the Ergowear test set, 
    (\textbf{b}) Model-free LSTM on the MTwAwinda test set, 
    (\textbf{c}) LSTM-MG-detach on the Ergowear test set, and 
    (\textbf{d}) LSTM-MG-detach on the MTwAwinda test set.}
    \label{fig:boxplots_NN}
\end{figure*}

Figure \ref{fig:boxplots_NN} presents the error per segment of model-free LSTM and LSTM-MG-detach, on both Ergowear and MTwAwinda. As observed, the error per segment of the model-free LSTM trained and tested on Ergowear (fig. \ref{fig:boxplots_NN}.a) is more consistent than the one resulting from MTwAwinda ( fig. \ref{fig:boxplots_NN}.b), meaning that the third quartile, in almost all segments, was bellow the 30º degrees error mark on Ergowear data. The main contributor for the high average error and standard deviation are the upper limbs (i.e., hands, lower arm, and upper arm) possibly due to the high range of motion of the shoulder joint and consequent movement possibilities for the upper limbs. Also, this high average error is consistent with the pattern observed in the boxplot presented in figure \ref{fig:methods_acc_boxplot}.a, in terms of acceleration range. In turn, for Ergowear's LSTM-MG-detach (\ref{fig:boxplots_NN}.c), the greatest source of error is the neck segment, while for MTwAwinda's LSTM-MG-detach (\ref{fig:boxplots_NN}.d) the right hand and the left lower and upper leg were the segments with higher error.  


\subsubsection{Comparison with classical sensor fusion filters}

As a baseline for the deep learning-based experiments, we computed the classical fusion filters \gls{QAD} metrics, on both datasets. Different variants were tested to provide a broader comparison base, namely, Mahony filter without magnetometer data, Madgwick filter with and without magnetometer data, Madgwick filter with magnetometer rejection band, and the Extended Kalman Filter(\gls{EKF}) with magnetometer data. Table \ref{tab:table_classicalfilter_results} depicts the achieved results. 

As expected, the lower errors were obtained on the MTw Awinda dataset, ranging from 7.76º ± 6.75º to 17.03º ± 17.79º. In specific, the lowest \gls{QAD} was achieved for the Madgwick with magnetometer band rejection heuristic, closely followed by the Madgwick with magnetometer (7.87º ± 6.76º). Comparing to the results achieved by the \gls{NN}-based methods, one can observe that despite none of the \gls{NN}-based model presented in table \ref{tab:table_NN_results} overcomes the best classical fusion filter, i.e., the Madgwick with magnetometer band rejection heuristic, the QAD resulting from the TGCN-MG-detach model was very close (7.96º ± 6.72º), as well as the LSTM-MG-detach (8.08º ± 6.67º).

\begin{table*}[t]
    \centering
    \caption{Average \gls{QAD}º  and inference time (ms) results, on Ergowear and MTwAwinda data, of Classical Sensor Fusion filters and best NN-based sensor fusion method.}
    \maxsizebox{\textwidth}{!}
    {
    \begin{tabular}{l |c | c | c | c}
    Experiment                      & \multicolumn{2}{c|}{Ergowear}                  & \multicolumn{2}{c}{MTwAwinda}   \\ \cline{2-5}
                                      & QAD°                   & Inference time (ms)   & QAD°                & Inference time (ms) \\ \hline
        Mahony (w/o mag)              &  20.73 $\pm$ 19.39     & 0.38 $\pm$ 0.01       & 12.35 $\pm$ 14.00    & 0.38 $\pm$ 0.01        \\
        Madgwick (w/o mag)            &  20.71 $\pm$ 20.34     & 0.35 $\pm$ 0.01       & 11.24 $\pm$ 13.25    & 0.36 $\pm$ 0.01        \\
        Madgwick (w/ mag)             &  23.23 $\pm$ 17.58     & 1.15 $\pm$ 0.03       & 7.87 $\pm$ 6.76      & 1.16 $\pm$ 0.02        \\
        Madgwick (w/ mag reject)      &  22.86 $\pm$ 17.50     & 1.15 $\pm$ 0.02       & 7.76 $\pm$ 6.75      & 1.17 $\pm$ 0.01        \\
        EKF (w/mag)                   &  27.48 $\pm$ 20.56     & 1.09 $\pm$ 0.02       & 10.17 $\pm$ 7.11     & 1.12 $\pm$ 0.01        \\  
        Integral                      &  26.74 $\pm$ 27.77     & 0.18 $\pm$ 0.00       & 17.031 $\pm$ 17.79   & 0.19 $\pm$ 0.00        \\ \hline
        LSTM-MG-detach (NN-based)     &  20.63 $\pm$ 16.45     & 1.64 $\pm$ 0.02       & 8.08 $\pm$ 6.67      & 1.63 $\pm$ 0.03        \\ 
        TGCN-MG-detach (NN-based)     &  21.89 ± 17.00         & 1.84 $\pm$ 0.02       & 7.96 ± 6.72          & 1.16 $\pm$ 0.02         \\ \hline
    \end{tabular}
    }
    \label{tab:table_classicalfilter_results}
\end{table*}

For Ergowear dataset the \gls{QAD} ranged between  20.71º ± 20.34º and 27.48 ± 20.56, with best and worst fusion filters corresponding to the Madgwick without mag and the \gls{EKF}  without magnetometer, respectively. With this dataset, the best result was slightly outperformed by the LSTM-MG-detach algorithm( 20.63º $\pm$ 16.45º). Also, comparing the standard deviation between both experiments, one can observe that the angle estimation achieved with the NN-based algorithms is more precise. Aditionally, when comparing the classical fusion filter that used the magnetometer data, whose \gls{QAD} ranged between 22.86º ± 17.50º and 27.48º ± 20.56º, with \gls{NN}-based models, it is possible to observe that almost all \gls{NN}-based models outperform the classical filters, with the exception of models based on the \gls{TGCN}. Since Ergowear enviroment was full of magnetic interferences, this could indicate that \gls{NN}-based fusion filter are less affected by them. 

The presence of the magnetometer also has impact on the inference times. As observed, in the classical sensor fusion filter, adding a magnetometer increased the inference time from 0.35 ms to 1.15 ms in the case of the Madgwick algorithm. Furthermore, by comparing the information in table \ref{tab:table_NN_results} with that in table \ref{tab:table_classicalfilter_results}, one can observe that the inference time of the Hybrid-CFF-\glspl{NN} is higher than the combined inference times of the individual models.

\subsection{Benchmark and Ablation Studies}

\begin{table*}[t]
    \centering
    \caption{Average \gls{QAD}º results of the ablation studies on Ergowear data. To ease the results analysis, performance improvements are in bold and underlined, slight performance decreases ($<$ 1º) are underlined, and  performance decreases are in normal text, all in relation to the baseline results.}
    {\renewcommand{\arraystretch}{1.15}
    \maxsizebox{\textwidth}{!}{

    \begin{tabular}{l | c | c | c | c | c}
        Ablation study          & LSTM                                    & COMP-LSTM                                       & TGCN                                      & TGCN-COMP                                 & LSTM-MG-detach      \\ \hline 
        Baseline                & \textbf{21.57 $\pm$ 14.97}                       & \textbf{21.58 $\pm$  18.89}                           & \textbf{29.20 $\pm$  21.43}                           & \textbf{32.59 $\pm$  23.60}                        & \textbf{20.63 $\pm$  16.45}  \\  \hline 
        no augmentation         & \underline{22.20 $\pm$ 15.39}    & \underline{21.70 $\pm$ 18.17}       & \underline{29.73 $\pm$ 22.25}         &  {34.85 $\pm$ 23.23}       & \textbf{\underline{20.54 $\pm$ 16.38}}  \\ 
        output (6D)             &  {23.04 $\pm$ 16.82}      &  {23.30 $\pm$ 21.44}         &  \textbf{\underline{23.06 $\pm$ 17.17}}          &  \textbf{\underline{25.18 $\pm$ 23.25}}         & \underline{21.21 $\pm$ 16.35}  \\
        no magnetometer         &  {36.97 $\pm$ 29.07}      &  {33.95 $\pm$ 27.79}         &  {41.15 $\pm$ 31.45}           &  {42.41 $\pm$ 30.65}         & -                   \\ 
        window (50)             &  \textbf{\underline{17.48 $\pm$ 11.01}}    &  {27.44 $\pm$ 29.47}          & \underline{30.08 $\pm$  24.76}         &  {46.56 $\pm$ 34.47}         &  {22.70 $\pm$ 16.16}  \\ 
        window (5)              &  {46.08 $\pm$  36.07}     &  {84.60 $\pm$ 43.00}          & -                                             & -                                         &  {23.46 $\pm$  17.42}  \\  
        window (2)              &  {69.76 $\pm$  31.47}     &  {84.75 $\pm$ 47.76}          & -                                             & -                                         &  {23.11 $\pm$ 16.16}  \\ 
        window (1)              &  {109.41 $\pm$ 41.29}     &  {95.80 $\pm$ 43.13}          & -                                             & -                                         &  {23.17 $\pm$ 17.64}  \\ 
        MSE                     &  {30.80 $\pm$  25.67}     &  {33.46 $\pm$ 31.66}          & -                                             & -                                         &  {23.53 $\pm$ 19.68}  \\ 
        \textit{qdist}                   & \underline{22.12 $\pm$  15.57}     &  \textbf{\underline{21.08 $\pm$ 17.64}}       & -                                             & -                                         & \underline{21.23 $\pm$  16.97}  \\ 
        \textit{RelQAD}                  &  {35.28 $\pm$  19.77}     &  \textbf{\underline{21.37 $\pm$ 17.72}}         & -                                             & -                                         & \underline{20.78 $\pm$ 16.17}  \\ \hline 
        
    \end{tabular}
    }
    }
    \label{tab:table_abl_ergowear}
\end{table*}

\begin{table*}[t]
    \centering
    \caption{Average \gls{QAD}º results of the ablation studies on MTwAwinda data performance improvements are in bold and underlined, slight performance decreases ($<$ 1º) are underlined, and  performance decreases are in normal text, all in relation to the baseline results.}
    {\renewcommand{\arraystretch}{1.17}
    \maxsizebox{\textwidth}{!}{
    \begin{tabular}{l | c | c | c | c | c }
        Ablation study      & LSTM                                   & COMP-LSTM                              & TGCN                                 & TGCN-COMP                             & LSTM-MG-detach \\ \hline 
        Baseline            & \textbf{27.97 $\pm$  24.05}         & \textbf{24.30 $\pm$  22.63}               & \textbf{37.21 $\pm$  34.13}            & \textbf{38.68 $\pm$  36.13}           & \textbf{8.08 $\pm$  6.67}  \\ \hline 
        no augmentation     & \textbf{\underline{27.65 $\pm$ 24.05}}  &\underline{25.08 $\pm$ 23.95}    & \textbf{\underline{35.20 $\pm$ 32.54}}   &  \textbf{\underline{36.00 $\pm$ 32.91}}   &  \textbf{\underline{8.05 $\pm$ 6.85}}  \\ 
        output (6D)         & {29.53 $\pm$ 26.90}                  &{31.4 $\pm$ 30.58}                        & 30.23 $\pm$ 26.64   &  \textbf{\underline{34.73 $\pm$ 31.42}}   &  \textbf{\underline{7.82 $\pm$ 6.76}}  \\ 
        no magnetometer     & {62.92 $\pm$ 39.69}                  &{52.72 $\pm$ 37.88}                        & 66.87 $\pm$ 40.04   &  {65.87 $\pm$ 41.60}    & -                 \\
        window (50)         &  \textbf{\underline{14.29 $\pm$ 12.17}}  &  {98.18 $\pm$ 56.96}                  &  \textbf{\underline{19.56 $\pm$ 17.20}}   &  \textbf{\underline{30.24 $\pm$ 25.83}}   &  \textbf{\underline{7.58 $\pm$ 6.72}}  \\ 
        window (5)          & \underline{28.33 $\pm$ 30.83}         & {63.00 $\pm$ 48.63}                      & -                                    & -                                     &  \textbf{\underline{7.80 $\pm$ 6.76}}  \\ 
        window (2)          & {36.26 $\pm$ 28.74}                  & {47.77 $\pm$ 36.35}                       & -                                    & -                                     &  \textbf{\underline{7.87 $\pm$ 6.76}}  \\ 
        window (1)          & {69.82 $\pm$ 40.09}                  & {64.64 $\pm$ 40.19}                       & -                                    & -                                     &  \textbf{\underline{7.93 $\pm$ 7.92}}  \\ 
        MSE                 &{46.53 $\pm$ 36.94}                  & {41.36 $\pm$ 35.59}                        & -                                    & -                                     & \underline{8.37 $\pm$ 7.00}  \\ 
        \textit{qdist}               & \underline{28.55 $\pm$ 24.24}      &  \textbf{\underline{23.86 $\pm$ 22.61}}     & -                                    & -                                     & \underline{8.15 $\pm$ 6.85}  \\ 
        \textit{RelQAD}              & {77.33 $\pm$ 45.35}                 & \textbf{\underline{24.11 $\pm$ 22.40}}     & -                                    & -                                     &  \textbf{\underline{7.97 $\pm$ 6.76}}  \\ \hline 
    \end{tabular}
    }
    }
    \label{tab:table_abl_mtwawinda}
\end{table*}

\label{sec:results_ablation}
Multiple ablation studies were conducted by removing certain components of the complete pipeline, to identify the contribution of each to the overall results.
To ease the analysis, the results were divided by dataset, i.e., Ergowear and MTwAwinda, which are summarized in Table \ref{tab:table_abl_ergowear} and Table \ref{tab:table_abl_mtwawinda}. For a question of space and organization, the tables that contain the inference time of these models were added to the appendix (Table \ref{tab:table_abl_mtwawinda_inference} and Table \ref{tab:table_abl_ergowear_inference}.The findings are next described. 

\subsubsection{Data augmentation and regularization}
\label{sec:results_ablation_aug}
To understand the effects of data augmentation for the \glspl{NN} training, we conducted a experiment where the methods were trained without any data augmentation and regularization.

Starting by the Ergowear dataset, and comparing with the results achieved with the methods trained with data augmentation and regularization (baseline), in general the non-use of these techniques resulted in a slight performance decrease,  with average \gls{QAD} increasing between 0.58 \% and 6.93\%,  for the model-free \gls{LSTM} and \gls{TGCN}, and hybrid-complementary(COMP) \gls{LSTM} and\gls{TGCN}. The exception was the LSTM-MG-detach algorithms, where the average \gls{QAD} reduced 0.47\%..

On the MTw Awinda data, in turn, in relation to the baseline, it is possible to observe that the non-use of augmentation resulted in a improvement in the \gls{QAD} metric, with a reduction between 0.45\% and 6.92\%. The exception was the complementary LSTM, where \gls{QAD} increased 3.22\%.

\subsubsection{Output Representation}
\label{sec:results_ablation_output_repr}
Most works use quaternions to represent 3D rotations. However, recently, 6D continuous representations have been proposed suggesting that they outperform the commonly used 4D representation. This ablation test aims to understand if this extra effort, in terms of computation \cite{Learning-rotations} and implementation, is worthwhile.

For Ergowear data, table \ref{tab:table_abl_ergowear} shows that using a 6D representation only resulted in a performance improvement of the \gls{TGCN} model, both in model-free and complementary approaches, with a QAD decrease of 21.07\% and 22.75\%, respectively. regarding the remaining algorithms, the use of a 6D representation showed no positive results (QAD increase between 2.81\% and 7.99\%)

For MTwAwinda data, using the 6D output representation resulted in slight improvement of the LSTM-MG-detach (0.03 QAD, reduction of 3.23 \%), and similarly to the ergowear data, the complementary \gls{TGCN}. The remaining did not benefit from the introduction of the 6D output representation ( QAD increase between 5.56\% and 29.35\%). Further, as it can be observed in Tables \ref{tab:table_abl_ergowear_inference} and \ref{tab:table_abl_mtwawinda_inference} in Appendix A, the inference time of the models using this output representation increased, being the highest between the ablations tests.

\subsubsection{Use of Magnetometer}
\label{sec:results_representation}
The magnetometer is a sensor typically affected by magnetic interferences, which can increase the pose estimation error. To understand the impact of the use of magnetometer data and if the \glspl{NN} can handle magnetic interferences, we conducted an ablation study where all the algorithms were trained without using the magnetometer data. 

In both datasets, the removal of the magnetometer sensors 's affect all models' performance negatively, with the \gls{QAD} increasing between 30.12\% and 71.38\% for ergoware dataset, and between 70.31\% and 124.91\% for MTwAwinda dataset.

\begin{figure}[ht]
    \centering
    \includegraphics[width=0.9\textwidth]{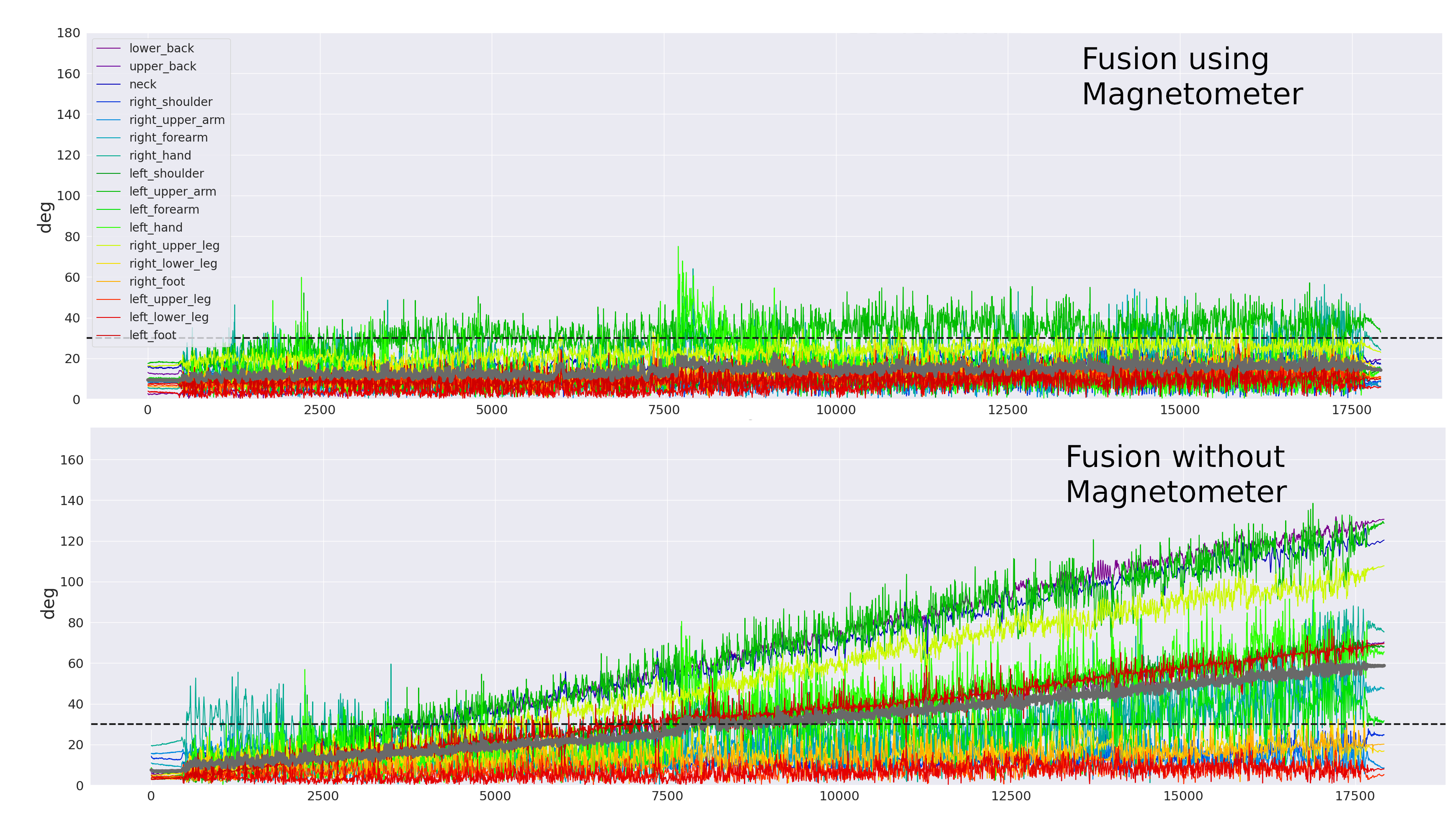}
    \caption{Temporal results on 5min trial with and without magnetometer, for the COMP-LSTM model.}
    \label{fig:results_ablation_5min_mag_nomag}
\end{figure}

\subsubsection{Use of Windows for training}
\label{sec:results_ablation_window}
To understand the trade-off of using window for training, a experiment was  conducted where \glspl{NN} were trained using windows of 50, 5, 2, and 1 samples. 

Starting the analysis by the ergowear dataset, from table \ref{tab:table_abl_ergowear} it is possible to observe that when using windows of 50 samples the \gls{LSTM} model outperformed considerably the baseline value. Also, when comparing to the classical sensor fusion filters from table \ref{tab:table_classicalfilter_results}, it is possible to observe that QAD obtained by the \gls{LSTM} trained with windows of 50 samples (17.48º ± 11.01º) is smaller than the one obtained by the Madgwick without magnetometer (20.71º ± 20.34º), which was the best method in terms of QAD. However, for the remaining \gls{NN} models, the effects of these 50 samples windows were not positive. For the smaller window sizes (5, 2, and 1 sample) the resulting \gls{QAD} value was considerably higher, being the highest achieved by 1 sample windows (109.41º ± 41.29º for LSTM, 407\% higher than the baseline). Notwithstanding, the LSTM-madgwick\textunderscore detach was not so affected by the window size as the other algorithms.

Regarding MTwAwinda dataset, similar to the Ergowear dataset, the use of 50 sample window led to the best results in terms of QAD, in almost all models, to the exception of the complementary \gls{LSTM}. In fact, the effect was very notorious on the \gls{LSTM} model, whose \gls{QAD} lowered from the baseline's 27.97º± 24.05º  to 14.29º ± 12.17º (reduction of 48.82\%) and on the LSTM-madgwick\textunderscore detach, which attained the best \gls{QAD} lowest \gls{QAD} of all the experiments, with 7.58º ± 6.76º, lower than the \gls{QAD} achieved with the Madgwick with magnetometer rejection (7.76º ± 6.75º). The use of small windows, with 1,2, and 5 samples, on the other hand, only seemed to benefit the LSTM-madgwick\textunderscore detach model, with all windows achieving a \gls{QAD} slightly inferior to the baseline. However, it is possible to observe that smaller the window size, higher the \gls{QAD}.

\subsubsection{Use of different loss functions}
\label{sec:results_ablation_lossfunctions}
These experiments were conducted to test the impact of different loss functions, namely MSE, \textit{qdist}, and \textit{RelQAD}, during the \gls{NN}s training. Overall, no significant \gls{QAD} decreases were verified. 

On Ergowear data, the only cases where \gls{QAD} decreased was with Complementary LSTM, trained with the \textit{qdist} (21.08º ± 17.64º, reduction of 2.30\%) and \textit{RelQAD} (21.37º ± 17.72º, reduction of 0.95\%)  loss functions,  The remaining experiments resulted in higher \gls{QAD}s, with increases between 0.73\% and 63.57\%.  The \gls{MSE} loss function was the

On the MTwAwinda data, the results pattern was similar, with \gls{QAD} being decreased when training the Complementary LSTM, with the \textit{qdist} (23.86º ± 22.61º, reduction of 1.81\%) and \textit{RelQAD} (24.11º ± 22.40º, reduction of 0.77\%) loss function, and, contrary to the Ergowear data, training the LSTM-madgwick\textunderscore detach with \textit{RelQAD} also resulted in a \gls{QAD} decrease (7.97º ± 6.76º, reduction of 1.37\%). The remaining experiments also resulted in higher QADs, with the increase ranging between 0.79\% and 176.42\%

 Aditionally, in both datasets (table \ref{tab:table_abl_ergowear} and \ref{tab:table_abl_mtwawinda}), it was possible to observe the \gls{MSE} loss function was the one that led to the worst performances, the \textit{RelQAD} loss fuction seems to be a bad fit for the model-free \gls{LSTM}, and the \textit{qdist} function seems to lead similar results to the \gls{QAD} as loss function.

\section{Discussion}
\label{sec:discussion}

Three different approaches were proposed to study the use of \glspl{NN}-based methods for Inertial pose estimation and benchmarked against commonly used algorithms, on data representative of low cost (Ergowear) and high end (Movella).

In general, the \gls{NN} based models were able to obtain comparative results to existing classical fusion-filter approaches, but not outperform them consistently.

Notwithstanding, the existence of a high amount of sensor-to-segment offsets made it harder to compare results. This indicates that when dealing with whole-body sensor fusion solutions, the complete processing pipeline needs to be considered, and not only the sensor fusion stage, in order to obtain the low errors needed for real-world applications.

The best results were, in general, achieved with the MTwAwinda dataset. This was expected given the higher quality data from the sensors and sensor-to-segment calibration method used.

Regarding the classical filters, these obtained relatively large errors, when compared to values in the literature, especially in the Ergowear dataset. This was mostly due to the presence of constant sensor-to-segment offsets. Nevertheless, these results served as baselines to compare novel approaches (Table \ref{tab:table_classicalfilter_results}).
The lower error obtained by the classical filters without using the data from the magnetometer can be attributed to the presence of magnetic disturbances and the relatively short trials. This is especially true for the Ergowear data, which was obtained in a more susceptible environment. Withal, the use of magnetometer data is necessary to prevent unbounded drift which accumulates over time, which is seen in longer trials (Figure \ref{fig:results_ablation_5min_mag_nomag}).

The magnetic rejection heuristic was able to improve results, compared to the default filter implementations with magnetometer, as also found by \cite{Madgwick2020ECF}. This indicates the presence of magnetic interference during the trials, and the attention needed to mitigate this issue. The \glspl{NN} were capable of somewhat dealing with the magnetic disturbances, obtaining better results than the baseline filters with magnetometer. The improvement was lesser in the MTwAwinda dataset, given the lower amount of disturbances, due to the added attention to this issue during acquisition.

Of the \gls{NN} backbone models tested, the \gls{LSTM} performed best in general, closely followed by \gls{GRU}, while the graph-based \gls{TGCN} performed much worse (Table \ref{tab:table_NN_results}).

The model-free \glspl{NN} were not able to consistently outperform the classical fusion filters. In the Ergowear datasets, the models obtained slightly better results, which might indicate the ability to compensate some of the lower quality data and calibration. However, on the MTwAwinda dataset, with better quality data, the models were far from outperforming the classical filters. Withal, the higher error obtained with this dataset (in comparison to ergowear) was not expected, but might be attributed to the higher complexity in fusion the full body sensor data (17 full-body vs 9 upper-body sensors) and lower number of training samples (2.5M vs 1M timesteps).

The complementary \glspl{NN} results were slightly better than the ones obtained with the model-free \gls{NN} approach. Still, they were not capable to surpass the classical fusion filters, even though this method relies on the gyroscope integration from the classical "Integral" algorithm. Given the higher effort in its implementation and instability during training (when using windows, as discussed bellow), this method should not be preferred (at least with this simplistic approach, to the model-free \gls{NN} approach).

The hybrid Madgwick filter and \glspl{NN}, was the one that presented the best results. On both Ergowear and MTwAwinda datasets the methods were able to achieve very similar results to the best classical sensor fusion filter for each dataset. In fact, for the Ergowear dataset, the method was capable of surpassing the classical method, although with a little difference. Even though, the \gls{QAD} error remained, in the best case, 20.63º for the Ergowear and 7.80º for the MTwAwinda datasets.

Regarding the ablation studies, starting by the data augmentation and regularization, the use of these methods did not seem to bring improvements, possibly given the large amount of data present in the datasets, the relatively small \gls{NN} models and the simple augmentation strategies used.

In terms of output representation, the 6D representation did not appear to improve results compared to the default quaternions, contrary to the findings of \cite{Zhou2019repr6d}. 
In relation to the use of magnetometer, as expected, it proved to have a considerable impact when using \glspl{NN}. This reinforces the need of magnetometer data and is indicative of the struggles with stable integration over long time horizons, while on the contrary being easier to fuse non-linear data from the accelerometer, magnetometer, and gyroscope to obtain the sensor orientation.

The windowed training made it difficult to optimize complementary-based models. This results from the drift and consequent error accumulation, which when using small window intervals becomes insignificant and thus ignored when optimizing, leading to high accumulation over the larger time-frames used for testing. Nevertheless, training with windows can and should be used when training non-complementary models, since this allows better train parallelization, speeding up training and lower prediction error. 
However, training with very small temporal windows ($<$5 samples), overall, led to worse results, making this an important parameter to consider.

As for the use of different loss functions, the traditionally used \gls{MSE} had the worst results. The \textit{qdist} performed much better, with results similar to \gls{QAD}. A relative root approach was also tried, yielding similar results for the complementary \gls{NN} approach, but much worse results for the model-free approach, which further evidences that model-free \gls{NN} mostly ignores the integration of orientation over a long time horizon.

The inference time of the filters used ranged from around 0.3ms (model-free \glspl{NN}) to 2ms (Hybrid-CFF-\glspl{NN}), on CPU (Intel\textsuperscript{TM} i9-10940X 3.30 GHz), with the the higher values being observed in the LSTM-MG-detach models. Thus, the deployment of the full-body fusion algorithms should not be an issue in most settings, since all of its components (classical filters, \glspl{RNN}) are suitable for real-time inference, and the computation can be vectorized for any number of sensors used.

\section{Conclusion}
\label{sec:conclusion}

The goal of this work was to study the use of methods based on \glspl{NN} for the estimation of inertial pose, with the aim of a method capable of abstracting from complex biomechanical models and analytical mathematics.

This work indicates that \glspl{NN} can be trained to perform full-body sensor fusion, with results comparable to classical filters. In addition, it suggests that a combination of classical filters with \glspl{NN} methods could lead to improvements with model-based principled outputs of the classical filters, with the data-driven adaptability of \glspl{NN} to deal with possible perturbations.  However, depending on the accuracy requirements of specific applications, the current model may require further refinement to meet those needs.

Nevertheless, solving the full-body inertial pose estimation problem requires not only addressing the errors arising from the sensor fusion step but also considering the entire pipeline, including sensor-to-segment calibration. Therefore, future work should explore more advanced sensor-to-segment calibration techniques. Additionally, given the complexity of the problem due to the high degrees of freedom in the human body, the existing dataset should be expanded to provide more data for re-training the models.


\section*{Acknowledgments}
\noindent This work is supported by: European Structural and Investment Funds in the FEDER component, through the Operational Competitiveness and Internationalization Programme (COMPETE 2020) [Project nº 39479; Funding Reference: POCI-01-0247-FEDER-39479]; national support to R\&D units grant through the reference project UIDB/04436/2020 and UIDP/04436/20200. Sara Cerqueira was supported by the doctoral Grant SFRH/BD/151382/2021, financed by the Portuguese Foundation for Science and Technology (FCT), under MIT Portugal Program.

\section*{Author contributions statement}

S.C., M.P. and C.P.S. contributed to the conceptualization and methodology design of the study ; M.P. developed the software; S.C. conducted validation and visualization. S.C., M.P., and C.P.S. analyzed the results; S.C. and M.P wrote the first draft of the manuscript. S.C., M.P. and C.P.S. contributed to manuscript revision and approved the submitted version. C.P.S. supervised the work and founded the research project.

\section*{Declaration of Competing Interests}
\noindent The authors declare that they have no known competing financial interests or personal relationships that could have appeared to influence the work reported in this paper.

\appendix
\section[\appendixname~\thesection]{}
\clearpage 
\begin{table*}[t]
    \centering
    \caption{Average \gls{QAD}º and Inference time (ms) results of the ablation studies on Ergowear data. To ease the results analysis, performance improvements are in bold and underlined, slight performance decreases ($<$ 1º) are underlined, and  performance decreases are in normal text, all in relation to the baseline results.}
    {\renewcommand{\arraystretch}{1.5}
    \maxsizebox{\textwidth}{!}{

    \begin{tabular}{l | c | c | c | c | c | c | c | c | c | c}
        Ablation study      & \multicolumn{2}{c|}{LSTM}                         & \multicolumn{2}{c|}{COMP-LSTM }                      & \multicolumn{2}{c|}{TGCN}                                   & \multicolumn{2}{c|}{TGCN-COMP}                       & \multicolumn{2}{c|}{LSTM-MG-detach}      \\ \hline 
                            & QAD (º)                    & Inference Time (ms)  & QAD (º)                       & Inference Time (ms)  & QAD (º)                              & Inference Time (ms)  & QAD (º)                        & Inference Time (ms)  & QAD (º)                               & Inference Time (ms)\\ \hline
        Baseline            &\textbf{21.57 $\pm$ 14.97}     &{0.32 $\pm$ 0.01}   & \textbf{21.58 $\pm$  18.89}   &{0.46 $\pm$ 0.01}     &\textbf{29.20 $\pm$  21.43}           & {0.50 $\pm$ 0.01}    & \textbf{32.59 $\pm$  23.60}    & {0.63 $\pm$ 0.01}  & \textbf{20.63 $\pm$  16.45}             & {1.64 $\pm$ 0.02}\\  \hline 
        no augmentation     &\underline{22.20 $\pm$ 15.39}  &{0.32 $\pm$ 0.01}  & \underline{21.70 $\pm$ 18.17} &{0.32 $\pm$ 0.01} &\underline{29.73 $\pm$ 22.25}             &{0.54 $\pm$ 0.01}     & {34.85 $\pm$ 23.23}            & {0.66 $\pm$ 0.01}   & \textbf{\underline{20.54 $\pm$ 16.38}} & {1.60 $\pm$ 0.02}\\ 
        output (6D)         &{23.04 $\pm$ 16.82}            &{0.72 $\pm$ 0.01}  &  {23.30 $\pm$ 21.44}          &{0.84 $\pm$ 0.01}     &\textbf{\underline{23.06 $\pm$ 17.17}}&{0.72 $\pm$ 0.01} &\textbf{\underline{25.18 $\pm$ 23.25}}&{0.85 $\pm$ 0.01}  & \underline{21.21 $\pm$ 16.35}          & {1.98 $\pm$ 0.02}\\
        no magnetometer     &{36.97 $\pm$ 29.07}            &{0.32 $\pm$ 0.01}  &  {33.95 $\pm$ 27.79}          &{0.48 $\pm$ 0.01}     & {41.15 $\pm$ 31.45}                  &{0.52 $\pm$ 0.01}     & {42.41 $\pm$ 30.65}            & {0.66 $\pm$ 0.01}   & -                                      & -  \\ 
        window (50) &\textbf{\underline{17.48 $\pm$ 11.01}} &{0.33 $\pm$ 0.01}  &  {27.44 $\pm$ 29.47}          &{0.33 $\pm$ 0.01}     & \underline{30.08 $\pm$  24.76}       &{0.70 $\pm$ 0.11}     & {46.56 $\pm$ 34.47}            & {0.81 $\pm$ 0.06}   & {22.70 $\pm$ 16.16}                    & {1.82 $\pm$ 0.06}\\ 
        window (5)          &{46.08 $\pm$  36.07}           &{0.33 $\pm$ 0.01}  &  {84.60 $\pm$ 43.00}          &{0.47 $\pm$ 0.01}     & -                                    & -                    & -                              & -                   & {23.46 $\pm$  17.42}                   & {1.59 $\pm$ 0.02}\\  
        window (2)          &{69.76 $\pm$  31.47}           &{0.32 $\pm$ 0.01}  &  {84.75 $\pm$ 47.76}          &{0.45 $\pm$ 0.01}     & -                                    & -                    & -                              & -                   & {23.11 $\pm$ 16.16}                    & {1.58 $\pm$ 0.02}\\ 
        window (1)          &{109.41 $\pm$ 41.29}           &{0.33 $\pm$ 0.01}  &  {95.80 $\pm$ 43.13}          &{0.46 $\pm$ 0.01}     & -                                    & -                    & -                              & -                   & {23.17 $\pm$ 17.64}                    & {1.61 $\pm$ 0.01}\\ 
        MSE                 &{30.80 $\pm$  25.67}           &{0.32 $\pm$ 0.01}  &  {33.46 $\pm$ 31.66}          &{0.32 $\pm$ 0.01}     & -                                    & -                    & -                              & -                   & {23.53 $\pm$ 19.68}                    & {1.59 $\pm$ 0.03}\\ 
        \textit{qdist}  &\underline{22.12 $\pm$  15.57}   &{0.32 $\pm$ 0.01}  &\textbf{\underline{21.08 $\pm$ 17.64}}  &{0.46 $\pm$ 0.01}  & -                                & -                    & -                              & -                   &\underline{21.23 $\pm$ 16.97}           & {1.61 $\pm$ 0.01}\\ 
        \textit{RelQAD}     &{35.28 $\pm$  19.77}           &{0.32 $\pm$ 0.04}      &  \textbf{\underline{21.37 $\pm$ 17.72}}    &{0.47 $\pm$ 0.01}      & -                           & -        & -                           & -                                                             & \underline{20.78 $\pm$ 16.17}          & {1.63 $\pm$ 0.02}\\ \hline 
    \end{tabular}
    }
    }
    \label{tab:table_abl_ergowear_inference}
\end{table*}

\begin{table*}[t]
    \centering
    \caption{Average \gls{QAD}º and inference time (ms) results of the ablation studies on MTwAwinda data performance improvements are in bold and underlined, slight performance decreases ($<$ 1º) are underlined, and  performance decreases are in normal text, all in relation to the baseline results.}
    {\renewcommand{\arraystretch}{1.5}
    \maxsizebox{\textwidth}{!}{
    \begin{tabular}{l | c | c | c | c | c | c | c | c | c | c}
        Ablation study     & \multicolumn{2}{c|}{LSTM}                                 & \multicolumn{2}{c|}{COMP-LSTM }                        & \multicolumn{2}{c|}{TGCN}                                  & \multicolumn{2}{c|}{TGCN-COMP}                              & \multicolumn{2}{c|}{LSTM-MG-detach}      \\ \hline 
                        & QAD (º)                             & Inference Time (ms)  & QAD (º)                            & Inference Time (ms) & QAD (º)                             & Inference Time (ms)   & QAD (º)                             & Inference Time (ms)  & QAD (º)                             & Inference Time (ms)\\ \hline
        Baseline        &\textbf{27.97 $\pm$  24.05}             &{0.34 $\pm$ 0.01}  & \textbf{24.30 $\pm$  22.63}           &{0.49 $\pm$ 0.01} & \textbf{37.21 $\pm$  34.13}            & {0.60 $\pm$ 0.01}  &\textbf{38.68 $\pm$ 36.13}             &{0.74 $\pm$ 0.02}    &\textbf{8.08 $\pm$  6.67}           &{1.63 $\pm$ 0.03}  \\ \hline 
        no augmentation &\textbf{\underline{27.65 $\pm$ 24.05}}  &{0.57 $\pm$ 0.01}  &\underline{25.08 $\pm$ 23.95}          &{0.72 $\pm$ 0.01} & \textbf{\underline{35.20 $\pm$ 32.54}} & {0.33 $\pm$ 0.01}  &\textbf{\underline{36.00 $\pm$ 32.91}} &{0.48 $\pm$ 0.01}   &\textbf{\underline{8.05 $\pm$ 6.85}} &{1.64 $\pm$ 0.02} \\ 
        output (6D)     & {29.53 $\pm$ 26.  90}                  &{0.84 $\pm$ 0.01}  &{31.4 $\pm$ 30.58}                     &{0.89 $\pm$ 0.21} & 30.23 $\pm$ 26.64                      & {0.79 $\pm$ 0.03}  &\textbf{\underline{34.73 $\pm$ 31.42}} &{0.89 $\pm$ 0.01}   &\textbf{\underline{7.82 $\pm$ 6.76}} &{0.32 $\pm$ 0.01} \\ 
        no magnetometer & {62.92 $\pm$ 39.69}                    &{0.36 $\pm$ 0.04}  &{52.72 $\pm$ 37.88}                    &{0.50 $\pm$ 0.01} & 66.87 $\pm$ 40.04                      & {0.59 $\pm$ 0.01}  &{65.87 $\pm$ 41.60}                    &{0.73 $\pm$ 0.01}   & -                                   & -   \\
        window (50)     &\textbf{\underline{14.29 $\pm$ 12.17}}  &{0.53 $\pm$ 0.04}  &{98.18 $\pm$ 56.96}                    &{0.37 $\pm$ 0.04} & \textbf{\underline{19.56 $\pm$ 17.20}} & {0.56 $\pm$ 0.01}  &\textbf{\underline{30.24 $\pm$ 25.83}} &{0.69 $\pm$ 0.01}   &\textbf{\underline{7.58 $\pm$ 6.72}} &{1.89 $\pm$ 0.05} \\ 
        window (5)      &\underline{28.33 $\pm$ 30.83}           &{0.34 $\pm$ 0.03}  &{63.00 $\pm$ 48.63}                    &{0.47 $\pm$ 0.01} & -                                      & -                  & -                                     & -                  &\textbf{\underline{7.80 $\pm$ 6.76}} &{1.84 $\pm$ 0.05} \\ 
        window (2)      & {36.26 $\pm$ 28.74}                    &{0.33 $\pm$ 0.01}  &{47.77 $\pm$ 36.35}                    &{0.47 $\pm$ 0.01} & -                                      & -                  & -                                     & -                  &\textbf{\underline{7.87 $\pm$ 6.76}} &{1.62 $\pm$ 0.01} \\ 
        window (1)      & {69.82 $\pm$ 40.09}                    &{0.34 $\pm$ 0.02}  &{64.64 $\pm$ 40.19}                    &{0.49 $\pm$ 0.01} & -                                      & -                  & -                                     & -                  &\textbf{\underline{7.93 $\pm$ 7.92}} &{1.66 $\pm$ 0.01} \\ 
        MSE             &{46.53 $\pm$ 36.94}                     &{0.36 $\pm$ 0.04}  &{41.36 $\pm$ 35.59}                    &{0.49 $\pm$ 0.01} & -                                      & -                  & -                                     & -                  &\underline{8.37 $\pm$ 7.00}          &{1.66 $\pm$ 0.02}  \\ 
        \textit{qdist}  &\underline{28.55 $\pm$ 24.24}           &{0.35 $\pm$ 0.02}  &\textbf{\underline{23.86 $\pm$ 22.61}} &{0.48 $\pm$ 0.01} & -                                      & -                  & -                                     & -                  &\underline{8.15 $\pm$ 6.85}          &{1.68 $\pm$ 0.02}  \\ 
        \textit{RelQAD} & {77.33 $\pm$ 45.35}                    &{0.35 $\pm$ 0.05}  &\textbf{\underline{24.11 $\pm$ 22.40}} &{0.49 $\pm$ 0.01} & -                                      & -                  & -                                     & -                  &\textbf{\underline{7.97 $\pm$ 6.76}} &{1.67 $\pm$ 0.06}  \\ \hline 
    \end{tabular}
    }
    }
    \label{tab:table_abl_mtwawinda_inference}
\end{table*}

\printglossaries

 \bibliography{cas-refs}





\end{document}